\pdfoutput=1
\RequirePackage{fix-cm}

\documentclass[twocolumn]{svjour3}           % twocolumn

\usepackage[T1]{fontenc}
\usepackage[utf8]{inputenc}
\usepackage[english]{babel}
\usepackage{newtxtext}
\usepackage[vvarbb]{newtxmath}
\usepackage{courier}
\usepackage{textcomp}
\usepackage{hyphenat}
\usepackage[all]{nowidow}
\usepackage{natbib}

\usepackage{amsmath}
\usepackage{amssymb}
\usepackage{amsfonts}
\usepackage{xfrac}

\usepackage{soul}  % colors, underlining etc.
\usepackage{lipsum}

\newcommand{\subparagraph}{} % Hack to make titlesec work
\usepackage{titlesec}

\usepackage{etoolbox}
\makeatletter
  \patchcmd{\ttlh@hang}{\parindent\z@}{\parindent\z@\leavevmode}{}{}
  \patchcmd{\ttlh@hang}{\noindent}{}{}{}
\makeatother

\usepackage{booktabs}
\usepackage{tabulary}
\usepackage[leftcaption]{sidecap}

\usepackage{epsfig}
\usepackage{graphicx}
\usepackage[labelsep=period,labelfont=bf,font={small}]{caption}
\usepackage{subcaption}
\usepackage{dblfloatfix} % fix for 2-column float placement
\usepackage[hang,flushmargin]{footmisc}

\usepackage{pifont}

\usepackage[dvipsnames]{xcolor}
\usepackage{soul}

\usepackage[colorlinks=true]{hyperref}

\newcommand{\eg}{\textit{e}.\textit{g}. }

\graphicspath{{./images/}}

\newcommand{\comment}[1]{\ignorespaces}

\newcommand{\datasetname}{\emph{QUVA Repetition}}
\newcommand{\ytsegments}{\emph{YTSegments}}
\newcommand{\datasetnamebold}{\textbf{QUVA Repetition}}
\newcommand{\ytsegmentsbold}{\textbf{YTSegments}}

\def\*#1{\mathbf{#1}}  % vector shortcut
\DeclareMathAlphabet\mathcalbf{OMS}{cmsy}{b}{n}
\newcommand{\nablabf}{\boldsymbol{\nabla}}

\newcommand{\grad}{\nablabf \*F}
\newcommand{\divergence}{\nablabf \boldsymbol{\cdot} \*F}
\newcommand{\curl}{\nablabf \boldsymbol{\times} \*F}

\newcommand{\gradd}{\nablabf \mathcalbf{F}}
\newcommand{\divergencee}{\nablabf \boldsymbol{\cdot} \mathcalbf{F}}
\newcommand{\curll}{\nablabf \boldsymbol{\times} \mathcalbf{F}}

\DeclareMathOperator{\Diag}{diag}
\DeclareMathOperator{\Trace}{trace}

\titleformat*{\section}{\large\bfseries}
\titleformat*{\subsection}{\normalsize\bfseries}
\titleformat*{\subsubsection}{\normalsize\itshape}
\titlespacing\section{0pt}{12pt plus 4pt minus 2pt}{6pt plus 2pt minus 2pt}
\titlespacing\subsection{0pt}{10pt plus 2pt minus 2pt}{6pt plus 2pt minus 2pt}
\titlespacing\subsubsection{0pt}{12pt plus 4pt minus 2pt}{4pt plus 2pt minus 2pt}

\addto\extrasenglish{}

\definecolor{highlightcolor}{RGB}{255,237,191}
\definecolor{textboxcolor}{RGB}{155,155,155}
\definecolor{paperboxcolor}{RGB}{204,148,73}
\definecolor{highlightred}{RGB}{224,103,103}

\journalname{arXiv preprint}

\hypersetup{pdfstartview={FitH}}
\hypersetup{pdfborder={0 0 0}}
\hypersetup{pdfpagemode={UseNone}}
\hypersetup{pdfcreator={Tom F.H. Runia}}
\hypersetup{pdfproducer={Tom F.H. Runia}}
\hypersetup{pdftitle={Repetition Estimation}}
\hypersetup{colorlinks=true}
\hypersetup{pdfsubject={Computer Vision}}
\hypersetup{pdfauthor={Tom F.H. Runia, Cees G.M. Snoek, Arnold W.M. Smeulders}}
\hypersetup{pdfkeywords={Computer Vision, Repetition Estimation, Counting, Periodicity, Wavelets, Motion, Optical Flow}}
\hypersetup{citecolor=blue}
\hypersetup{linkcolor=blue}

\begin{document}

\title{Repetition Estimation}
%\title{Repetition Estimation }
%\title{Frequency-Based Repetition Counting in Video}
%\subtitle{\normalfont{Repetition estimation and action classification}}

\author{Tom F.H. Runia \and
        Cees G.M. Snoek \and
        Arnold W.M. Smeulders
}

\institute{QUVA Deep Vision Lab, University of Amsterdam \\
            Science Park 904, 1098XH Amsterdam, The Netherlands \\
           Corresponding Author: Tom F.H. Runia, \texttt{runia@uva.nl}
}

\date{Received: date / Accepted: date}

\maketitle

% See: 'how to write Nature abstract.jpg'
\begin{abstract}
  Visual repetition is ubiquitous in our world. It appears in human activity (sports, cooking), animal behavior (a bee's waggle dance), natural phenomena (leaves in the wind) and in urban environments (flashing lights). Estimating visual repetition from realistic video is challenging as periodic motion is rarely perfectly \emph{static} and \emph{stationary}. To better deal with realistic video, we elevate the static and stationary assumptions often made by existing work. Our spatiotemporal filtering approach, established on the theory of periodic motion, effectively handles a wide variety of appearances and requires no learning. Starting from motion in 3D we derive three periodic motion types by decomposition of the motion field into its fundamental components. In addition, three temporal motion continuities emerge from the field's temporal dynamics. For the 2D perception of 3D motion we consider the viewpoint relative to the motion; what follows are $18$ cases of recurrent motion perception. To estimate repetition under all circumstances, our theory implies constructing a mixture of differential motion maps: $\*{F}$, $\grad$, $\divergence$ and $\curl$. We temporally convolve the motion maps with wavelet filters to estimate repetitive dynamics. Our method is able to spatially segment repetitive motion directly from the temporal filter responses densely computed over the motion maps. For experimental verification of our claims, we use our novel dataset for repetition estimation, better-reflecting reality with non-static and non-stationary repetitive motion. On the task of repetition counting, we obtain favorable results compared to a deep learning alternative.

  % Please provide 4 to 6 keywords which can be used for indexing purposes.

\end{abstract}

\section{Introduction}
\label{sec:introduction}

% General introduction, appearances of visual repetition
Visual repetitive motion is common in our everyday experience as it appears in sports, music-making, cooking and other daily activities. In natural scenes, it appears as leaves in the wind, waves in the sea or the drumming of a woodpecker, whereas our encounters of visual repetition in urban environments include blinking lights, the spinning of wind turbines or a waving pedestrian. In this work we reconsider the theory of periodic motion and propose a method for estimating repetition in real-world video.

Improving our ability to estimate repetition in realistic video is important in numerous aspects. In computer vision, periodic motion has proven to be useful for action classification \citep{goldenberg2005behavior,lu2004repetitive}, action localization \citep{laptev2005periodic,sarel2005separating}, human motion analysis \citep{albu2008generic,ran2007pedestrian}, 3D reconstruction \citep{belongie2006structure}, animal behavior study \citep{davis2000categorical} and camera calibration \citep{huang2016camera}. From a biological perspective, repetition is fascinating as the human visual system relies on rhythm and periodicity to approximate velocity, estimate progress and to trigger attention \citep{johansson1973visual}.

% Repetition estimation can also serve as method for deriving velocity or traveled distance for humans walking or cycling.

\begin{figure*}
  \centering
  \includegraphics[width=\textwidth]{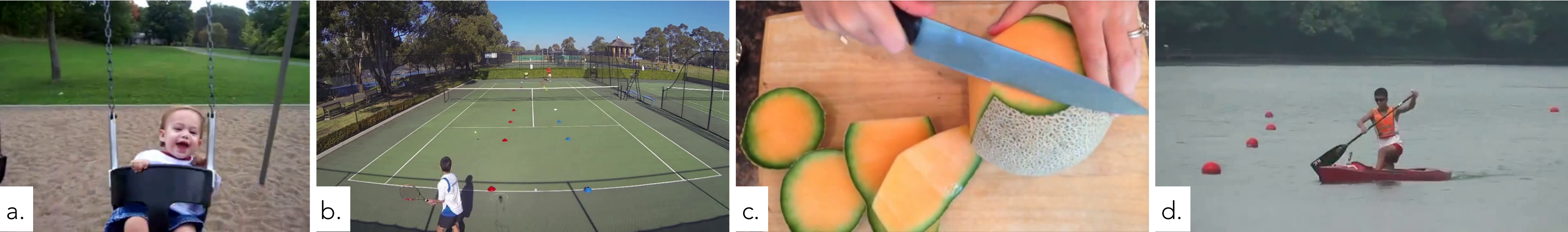}
  \caption{Examples from our \datasetname{} dataset, containing videos with repetitive motion such as sports, cooking, music making and other daily activities. The videos are challenging in their variety of appearance, non-stationary motion (\eg accelerations or transient phenomena) and non-static appearance induced by camera motion or a changing motion appearance throughout the video. In this paper we focus on dealing with such challenges as they often appear in the real-world.}
  \label{fig:intro-dataset-examples}
  \vspace{-1em}
\end{figure*}

% Introduces our theoretical contribution
To understand the origin and appearance of visual repetition we rethink the theory of periodic motion inspired by existing work \citep{pogalin2008visual,davis2000categorical}. We follow a differential geometric approach, starting from the divergence, gradient and curl components of the 3D flow field. From the decomposition of the motion field and its temporal dynamics, we derive three \emph{motion types} and three \emph{motion continuities} to arrive at $3\times3$ fundamental cases of intrinsic periodicity in 3D. For the 2D perception of 3D intrinsic periodicity, the observer's viewpoint can be somewhere in the continuous range between two viewpoint extremes. Finally, we arrive at $18$ fundamental cases for the 2D perception of 3D intrinsic periodic motion.

% Applications of visual repetitions
%Since the early days of computer vision, significant attention has been devoted to the concept of visual repetition, however 
Estimating repetition in practice remains challenging. First and foremost, repetition appears in many forms due to its diversity motion types and motion continuity (\autoref{fig:intro-dataset-examples}). Sources of variation in motion appearance include the action class, origin of motion and the observer's viewpoint. Moreover, the motion appearance is often \emph{non-static} due to a moving camera or as the observed phenomena develops over time. In practice, repetitions are rarely perfectly periodic but rather are \emph{non-stationarity}. Existing literature \citep{levy2015live,pogalin2008visual} generally assumes static and stationary repetitive motion. As reality is more complex, we here address the challenges involved with non-static and non-stationary by proposing a novel method for estimating repetition in real-world video.

% Introduces our method contribution
To deal with the diverse and possibly non-static motion appearance in realistic video, our theory implies representing the video with a mixture of first-order differential motion maps. For non-stationary temporal dynamics the fixed-period Fourier transform \citep{cutler2000robust,pogalin2008visual} is not suitable. Instead, we handle complex temporal dynamics by decomposing the motion into a time-frequency distribution using the continuous wavelet transform. To increase robustness and to be able to handle camera motion, we combine the wavelet power of all motion representations. Finally, we alleviate the need for explicit tracking \citep{pogalin2008visual} or motion segmentation \citep{runia2018real} by segmenting repetitive motion directly from the wavelet power. On the task of repetition counting, our method performs well on an existing video dataset and our novel \datasetname{} dataset which emphasizes on more realistic video.

%CS: deze extra header is volgens mij niet nodig, hangt er nu beetje los bij.
%\subsection{Main Contributions}
%\label{sec:intro-main-contributions}

A preliminary version of this work appeared as \citep{runia2018real}. The current manuscript largely maintains the original theory while making significant improvements to the method for repetition estimation. Specifically, we simplify our approach by removing the need for explicit motion segmentation prior to repetition estimation. Instead, we obtain a foreground motion segmentation directly from the wavelet filter responses densely computed over the motion maps. As the most discriminative motion representation is not known \emph{a priori}, our previous work employed a self-quality assessment to select the representation best measurable. However, selecting a single most discriminative representation is inherently unsuitable for handling significant variations due to camera motion or motion evolution over the course of the video. We improve this by combining the wavelet power of all representations for robustness and viewpoint invariance. Together the two improvements simplify our method while improving or giving comparable results on the task of repetition counting. More precisely, the contributions of our work are as follows:

\begin{itemize}
  \item We rethink the theory of periodic motion to arrive at a classification of periodic motion. Starting from the 3D motion field induced by an object periodically moving through space, we decompose the motion into three elementary components: divergence, curl and shear. From the motion field decomposition and the field's temporal dynamics, we identify $9$ fundamental cases of periodic motion in 3D. For the 2D perception of 3D periodic motion we consider the observer's viewpoint relative to the motion. Two viewpoint extremes are identified, from which $18$ cases of 2D repetitive appearance emerge.

  \item Our spatiotemporal filtering method addresses the wide variety of repetitive appearances and effectively handles non-stationary motion. Specifically, diversity in motion appearance handled by representing video as six differential motion maps that emerge from the theory. To identify the repetitive dynamics in the possibly non-stationary video, we use the continuous wavelet transform to produce a time-frequency distribution densely over the video. Directly from the wavelet responses we localize the repetitive motion and determine the repetitive contents. 
  %CS: Door deze implementatie statement hier te zetten suggereer je dat er een experiment komt. Ik zou het tussen neus en lippen door noemen verderop in paper.
  %Our novel implementation performs all filtering on the GPU resulting in large speed-up.

  \item Extending beyond the video dataset of \cite{levy2015live}, we propose a new dataset for repetition estimation, that is more realistic and challenging in terms of non-static and non-stationary videos. To encourage further research on video repetition, we will make the dataset and source code available as download. %\footnote{\url{http://tomrunia.github.io/projects/repetition/}}. 

\end{itemize}

\noindent The paper proceeds as follows: in Section~\ref{sec:related-work}, we provide an overview of related work. Section~\ref{sec:theory-of-repetition} introduces new theory on periodic motion to arrive at a classification of fundamental motion types and their appearance in video. Our theoretical insights are at the core of our method for repetition estimation, which is presented in Section~\ref{sec:methods}. The experiments in Section~\ref{sec:experiments} evaluate our method on two challenging video datasets. Section~\ref{sec:conclusion} summarizes and concludes the manuscript.

%!TEX root = ms.tex

\section{Related Work}
\label{sec:related-work}

%We consider the related literature on repetition estimation and summarize existing work on the categorization of motion types as it is relevant to our theory on periodic motion.

\subsection{Repetition Estimation}
\label{subsec:relwork-repetition}

Existing approaches for repetition estimation in video typically represent video as one-dimensional signals that preserve the repetitive structure of the motion. Subsequently, frequency information is often extracted by Fourier analysis \citep{azy2008segmentation,cutler2000robust,pogalin2008visual,tsai1994cyclic}, peak detection \citep{thangali2005periodic}, singular value decomposition \citep{chetverikov2006motion} or computational topology \citep{tralie2018quasi}. In general, the existing methods perform well under the assumptions of static and stationary videos. 

\begin{figure}
 \centering
 \includegraphics[width=\columnwidth,trim={0 0 0 0},clip]{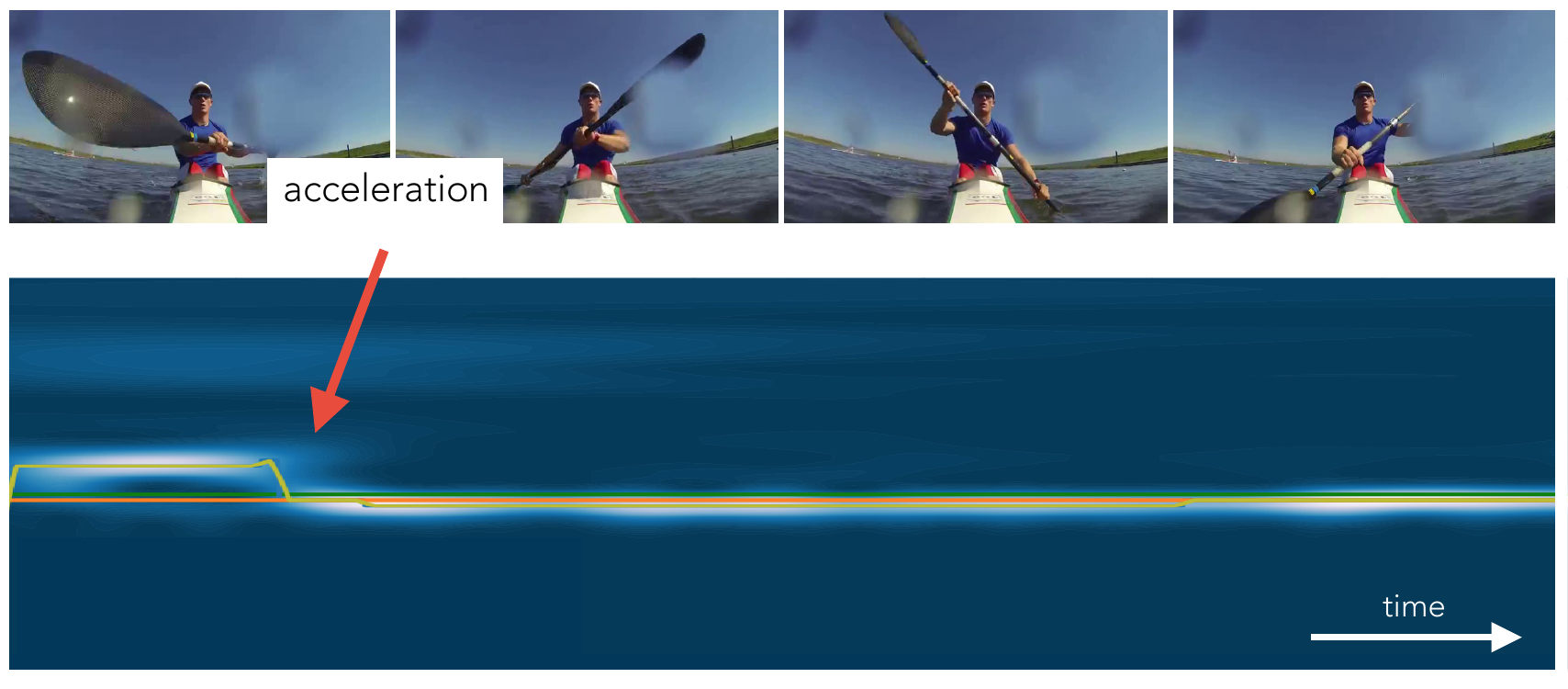}
 \vspace{-0.5cm}
 \caption{Non-stationary motion often appears in real-world video. This example shows a rower accelerating as plotted in the time-frequency space. The vertical axis of the spectrum denotes the wavelet scale, inversely proportional to the frequency. The sudden acceleration appears as shift of the maximum power in time-frequency space. The Fourier transform is unable to handle such non-stationary video. \label{fig:rower-example-nonstationary}}
 %\vspace{-0.4cm}
\end{figure}

The seminal work of \cite{cutler2000robust} uses normalized autocorrelation to obtain similarity matrices and proceeds by repetition estimation using Fourier analysis. \cite{pogalin2008visual} estimate the frequency of motion in video by tracking an object, performing principal component analysis over the tracked regions and also employing the Fourier-based periodogram. From the spectral decomposition, the dominant frequencies can be identified by peak detection and non-trivial separation of fundamental and harmonic frequencies. While Fourier-based methods provide a good estimate of strongly periodic motion, they are unsuitable nor intended to deal with more realistic non-stationary repetition, see the accelerating rower in \autoref{fig:rower-example-nonstationary}.

% However, to handle non-stationary motion as is ubiquitous in the real world, methods relying on Fourier-analysis for periodic motion are unsuitable nor intended.

% Paragraph on use of Fourier transform
% Most existing work on repetition estimation in video can be considered as similarity-based \citep{cutler2000robust}, filtering-based \citep{pogalin2008visual,burghouts2006quasi} or matching-based \citep{laptev2005periodic,sarel2005separating}. \todo{transition} More recently, \cite{levy2015live} proposes the use deep learning for periodic motion estimation and \cite{tralie2018quasi} relies on computational topology to quantify the level of periodicity.

% Similarity-based approaches generally adopt an inter-frame similarity or correlation measure to construct low-dimensional signals that act as a surrogate for the video dynamics.  More recently, \cite{tralie2018quasi} also constructs self-similarity matrices and computational topology for quantifying the level of periodicity. In general, similarity-based methods show good results under the assumption of static and clean video with challenging background clutter or repetition superposed on translation.

% Once a low-dimensional signal has been distilled to describe the video dynamics, the Fourier transform is often preferred for decomposing the signal's spectral components \citep{tsai1994cyclic,thangali2005periodic,pogalin2008visual}. 

% Paragraph on non-stationary analysis
As strongly periodic motion has received serious attention, less effort has been devoted to non-stationary repetition in video. \cite{briassouli2007extraction} use the Short-Time Fourier Transform for estimating the time-varying spectral components in video to distinguish multiple periodically moving objects. The filtering-based approach of \cite{burghouts2006quasi} uses a time-causal filter bank from \cite{koenderink1988scale} to detect quasi-periodic motion in video. Their method works online and shows good results when filter response frequencies are tuned correctly. In this work, we employ the continuous wavelet transform over multiple temporal scales to estimate repetition in complex video.

% Paragraph on Levy & Wolf
The deep learning method of \cite{levy2015live} is different from all other work but resembles our work in counting-based evaluation over a large video dataset. The general idea is to train a convolutional neural network for predicting the motion period in short video clips. As training data is not available, the network is optimized on synthetic video sequences in which moving squares exhibit periodic motion of four motion types from \cite{pogalin2008visual}. At test time, the method takes a stack of video frames, performs explicit motion localization to obtain a region of interest and then classifies the motion period by forwarding the frame crops through the network. The system is evaluated on the task of repetition counting and shows near-perfect performance on their \ytsegments{} dataset. The $100$ videos are a good initial benchmark but as the majority of videos have a static viewpoint and exhibit stationary periodic motion, we propose a new dataset. Our dataset better reflects reality by including more non-static and non-stationary examples. %Similar to \cite{levy2015live}, we also evaluate on the task of repetition counting.

% Visual apparent motion => Camera Motion, Segmentation
Increased video complexity in terms of motion appearance, scene complexity and camera motion demands intricate spatiotemporal localization of salient motion. While many methods for periodic motion analysis incorporate some form of tracking or motion segmentation \citep{polana1997detection,pogalin2008visual,levy2015live}, few approaches specifically address the challenge of repetitive motion segmentation. \cite{goldenberg2005behavior} estimate the repetitive foreground motion to leverages its center-of-mass trajectory for classifying human behavior. More closely related is the work of \cite{lindeberg2017dense} in which scale selection over space and time leads to an effective temporal scale map. Inspired by this, we perform spatial segmentation of repetitive motion directly from the spectral power maps obtained through the continuous wavelet transform. This is appealing, as it connects localization to the temporal dynamics rather than relying on decoupled localization by state-of-the-art motion segmentation, \eg \cite{tokmakov2017}.

% Applications of Periodic Motion Detection
Instead of considering repetition as their primary goal, various works leverage the presence of periodic motion for auxiliary tasks. \cite{belongie2006structure} exploit periodic human motion for 3D reconstruction of a scene, whereas \cite{laptev2005periodic} uses sequence alignment of periodic motion for stereo-camera correspondence. From a practical point of view, the presence of periodic motion also serves as cue for action classification \citep{lu2004repetitive,goldenberg2005behavior} and supports camera calibration \citep{huang2016camera}.

%%%%%%%%%%%%%%%%%%%%%%%%%%%%%%%%%%%%%%%%%%%%%%%%%%%%%%%%%%%%%%%%%%%%%%%%%%%%%%%%
%%%%%%%%%%%%%%%%%%%%%%%%%%%%%%%%%%%%%%%%%%%%%%%%%%%%%%%%%%%%%%%%%%%%%%%%%%%%%%%%
%%%%%%%%%%%%%%%%%%%%%%%%%%%%%%%%%%%%%%%%%%%%%%%%%%%%%%%%%%%%%%%%%%%%%%%%%%%%%%%%
% Motion Types

\begin{figure}
 \centering
 \includegraphics[width=\columnwidth,trim={0 0 0 0},clip]{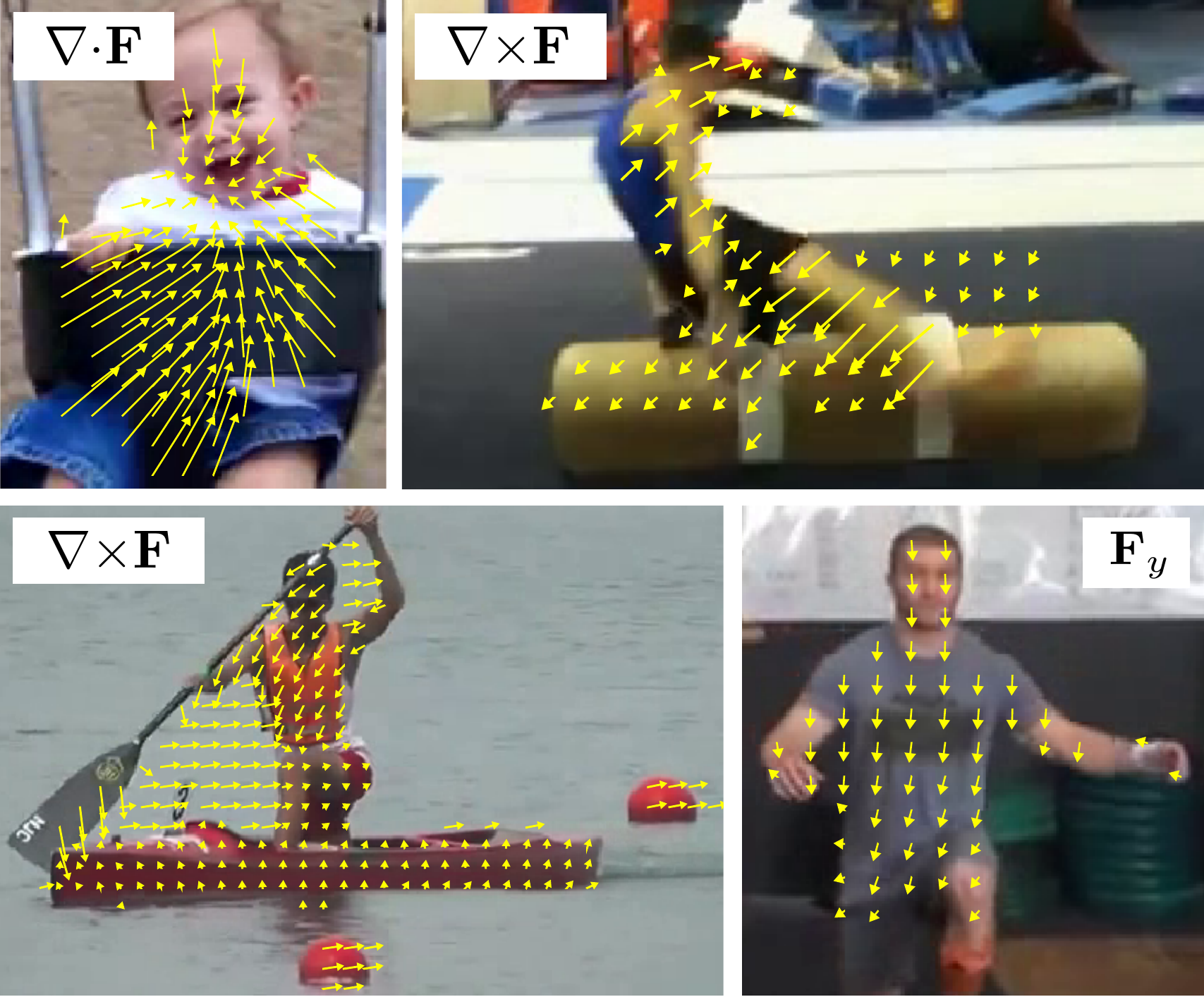}
 \caption{There is great diversity in appearance of repetitive motion. We decompose the motion field into its fundamental components. Here we visualize the motion fields as optical flow arrows over the foreground motion with the visually dominant motion field component indicated in the white box.
 \label{fig:flow-field-arrows}}
\end{figure}

\subsection{Categorization of Motion Types}
\label{subsec:relwork-motion-and-optical-flow}

% Paragraph on Koenderink
In real-world video, periodic motion emerges in a wide variety of appearances (see \autoref{fig:flow-field-arrows}). We reconsider the theory of periodic motion by proposing a classification of fundamental periodic motion types starting from the 3D motion field tied to a moving object. Using first-order differential analysis, we decompose the motion field into its primitive components. The work of \cite{koenderink1975invariant} delivered inspiration for our theoretical derivation of repetitive motion types from the flow field. Similar to the Helmholtz-Hodge decomposition \citep{abraham2012manifolds} into the eigenvalues of the flow field's Jacobian matrix, it finds use in flow field topology for fluid dynamics and electrodynamics. Although our work is similar in differential decomposition of the motion field, we use it to reach a novel classification of periodic motion patterns. We use the insights for establishing our repetition estimation method.

% Paragraph on Pogalin & Smeulders
In the context of periodic motion, \cite{davis2000categorical} and \cite{pogalin2008visual} both propose a categorization of motion patterns. \cite{davis2000categorical} consider a simple sinusoidal model to characterize periodic motion and link  each type to animal behavior. In terms of periodic motion categorization, our work bears resemblance to \cite{pogalin2008visual}. The authors identify four visually periodic motion types (translation, rotation, deformation and intensity variation) complemented with three cases of motion continuity (oscillating, constant and intermittent) in the field of view. We take a more principled approach starting from the 3D motion field. Specifically, we show that fundamental periodic motion types emerge from the decomposition of the flow field and the motion direction over time. Moreover, the \mbox{projection} of 3D periodicity on on a 2D image has to take into account the continuous nature of the viewpoint which we address explicitly in theory and experiments.

\begin{figure*}[ht]
  \centering
  \includegraphics[width=\textwidth]{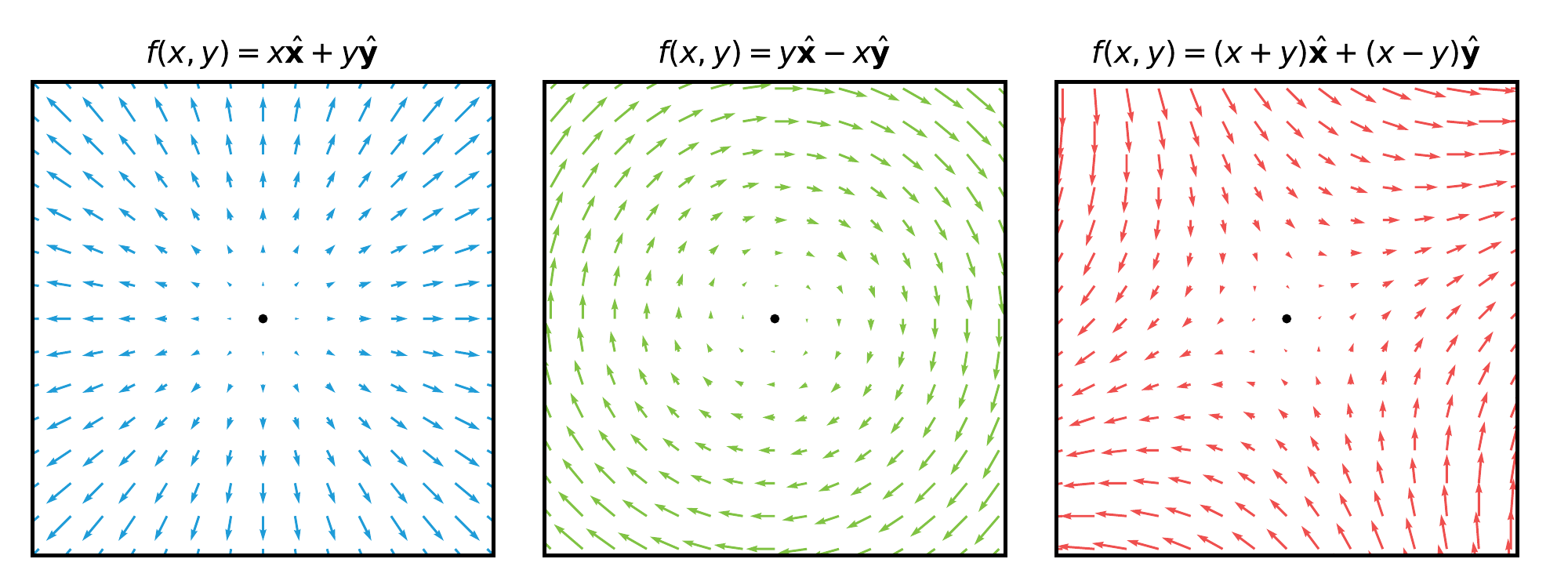}
  \caption{Three 2D flow fields with fundamentally different characteristics that emerge from the decomposition of the motion Jacobian $\*J$. \emph{Left}: Pure divergent flow field with outward flux often associated with expansion or depth perception. \emph{Center}: Pure rotational flow field also referred to as vorticity or curl. \emph{Right}: Flow field with a pure shear component related to the deformation tensor. The shear component is divergence- and curl-free as the opposing terms cancel out. In real-world video, shear is generally negligible compared to divergence and curl components. \label{fig:quiver-plot-all}}
\end{figure*}

Although not directly related to our work, first-order differential geometric motion representations have been used extensively as spatiotemporal video descriptors. \cite{klaser2008spatio} proposes a spatial multi-scale motion descriptor based on first-order differential motion and uses integral videos for  efficient computation. Along similar lines, MoSIFT \citep{chen2009mosift} uses spatial interest points and enforces sufficient temporal dynamics to eliminate candidate points. In terms of motion descriptors, our work bears resemblance to the Divergence-Curl-Shear descriptor proposed by \cite{jain2013better}. Their favorable action classification results associated with the differential-based descriptor support our findings for periodic motion estimation.

% Our theoretical contribution shares most similarity with \citep{koenderink1975invariant} and \citep{pogalin2008visual} as these works also consider classification of different motion types. From experimental perspective, our counting-based evaluation closely resembles \cite{levy2015live} although our method is orthogonal.

%%%%%%%%%%%%%%%%%%%%%%%%%%%%%%%%%%%%%%%%%%%%%%%%%%%%%%%%%%%%%%%%%%%%%%%%%%%%%%%%

%!TEX root = ms.tex

\section{Repetitive Motion}
\label{sec:theory-of-repetition}

Visual repetition is defined as a reoccurring pattern over space or time in the 3D world. In this work, we focus on temporally repetitive motion rather than spatially repetitive patterns such as a texture. Consequently, the 3D motion field induced by a moving object is the right starting point for our theoretical analysis.

Let a moving object and observer be positioned in a 3D world specified by the Cartesian coordinates $\*x = (x_1, x_2, x_3)$ at time $t$. Formally, intrinsic periodic motion is defined as the reappearance of the same 3D-flow $\mathcalbf{F}(\*x,t)$ induced by the motion of an object over time.
\begin{align}
    \mathcalbf{F}(\*x, t) = \mathcalbf{F}(\*x + \*S, t + T).
\end{align}
The parameter $T$ denotes the period over time and $\*S$ corresponds to a period over space. We initially exclude the trivial case of a constant flow field inducing periodic appearance due to a reappearing texture on the object's surface. Starting from the motion field, we follow a differential approach to decompose the field into its elementary components. In the end we arrive at nine fundamental cases of intrinsic periodic motion in 3D.

\subsection{Motion Field Decomposition}
\label{subsec:differential-motion-analysis}

In 3D Cartesian space, the gradient of the flow $\nablabf \mathcalbf{F}(\*x,t)$ is described by the Jacobian matrix $\*J \in \mathbb{R}^{3\times3}$ containing all first-order partial derivatives of the vector field:
\begin{align}
  (\nablabf \mathcalbf{F})_{ij} = \frac{\partial \mathcal{F}_i}{\partial x_j},
\end{align}
% %
% \begin{align}
%   \*\nabla \mathcalbf{F}(t) =
%   \begingroup % keep the change local
%   \setlength\arraycolsep{5pt}
%   \begin{bmatrix}
%     \frac{\partial \mathcal{F}_x}{\partial x} & \frac{\partial \mathcal{F}_y}{\partial x} & \frac{\partial \mathcal{F}_z}{\partial x} \\[7pt]
%     \frac{\partial \mathcal{F}_x}{\partial y} & \frac{\partial \mathcal{F}_y}{\partial y} & \frac{\partial \mathcal{F}_z}{\partial y} \\[7pt]
%     \frac{\partial \mathcal{F}_x}{\partial z} & \frac{\partial \mathcal{F}_y}{\partial z} & \frac{\partial \mathcal{F}_z}{\partial z}
%   \end{bmatrix}
%   \endgroup
% \end{align}
% %
% https://math.stackexchange.com/questions/1196427/what-does-shear-mean/
where $i$ and $j$ are dimension indices and we omit the position $\*x$ and time $t$ for brevity. From the first-order partial derivatives contained in the Jacobian, three fundamental components of the motion field can be recognized \citep{abraham2012manifolds}. Specifically, the Jacobian $\*J$ can be decomposed into a sum of a diagonal part $\*{D}$, a symmetric part $\*E$ and an anti-symmetric part $\*R$ such that:
\begin{align}
\nablabf \mathcalbf{F} = \*D + \*R + \*E.
\end{align}
This is similar to the Helmholtz-Hodge vector field decomposition well-known from fluid dynamics, which distinguishes divergence-free and curl-free components of a motion field. The diagonal part of the Jacobian $\*J$ is given by:
\begin{align}
  \*D = \Diag \left( \frac{\partial \mathcal{F}_1}{\partial x_1},\, \frac{\partial \mathcal{F}_2}{\partial x_2},\, \frac{\partial \mathcal{F}_3}{\partial x_3} \right).
  % \*D =
  % \begingroup % keep the change local
  % \setlength\arraycolsep{4pt}
  % \begin{bmatrix}
  %   \frac{\partial \mathcal{F}_x}{\partial x} & 0 & 0 \\[6pt]
  %   0 & \frac{\partial \mathcal{F}_y}{\partial y} & 0 \\[6pt]
  %   0 & 0 & \frac{\partial \mathcal{F}_z}{\partial z}
  % \end{bmatrix}.
  % \endgroup
\end{align}
The trace of this matrix defines the \emph{divergence} of the field:
\begin{align}
  \divergencee = \Trace \, (\*D). \label{eq:divergence-definition}
\end{align}

\begin{figure*}
\centering
    \begin{subfigure}[b]{0.45\textwidth}
        \includegraphics[width=\textwidth,trim={0 7.6cm 17cm 0},clip]{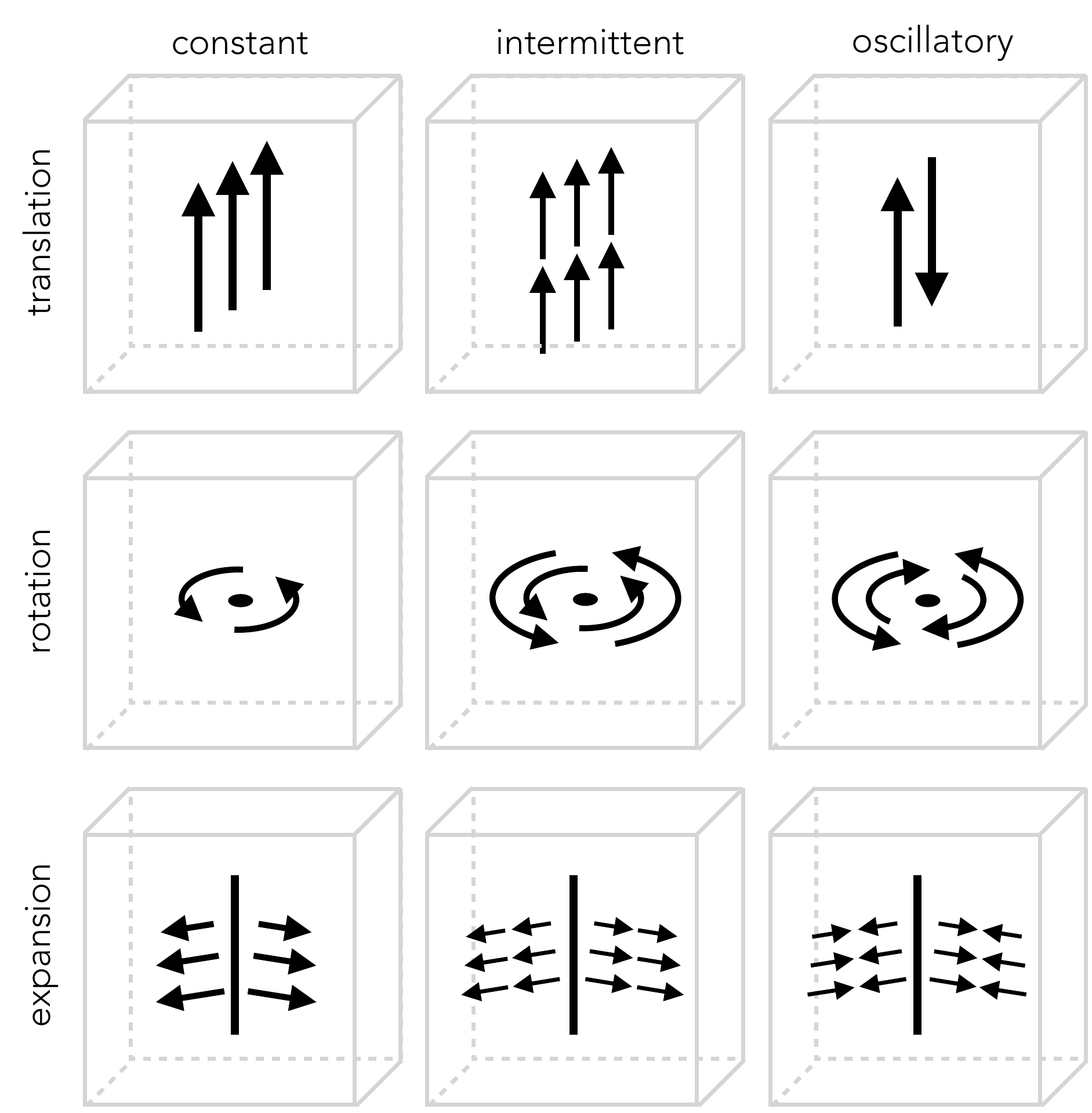}
        \caption{Flow Abstractions in 3D}
        \label{fig:3x3_cubes}
    \end{subfigure}
    \quad
    \begin{subfigure}[b]{0.45\textwidth}
        \includegraphics[width=\textwidth,trim={0 7.6cm 17cm 0},clip]{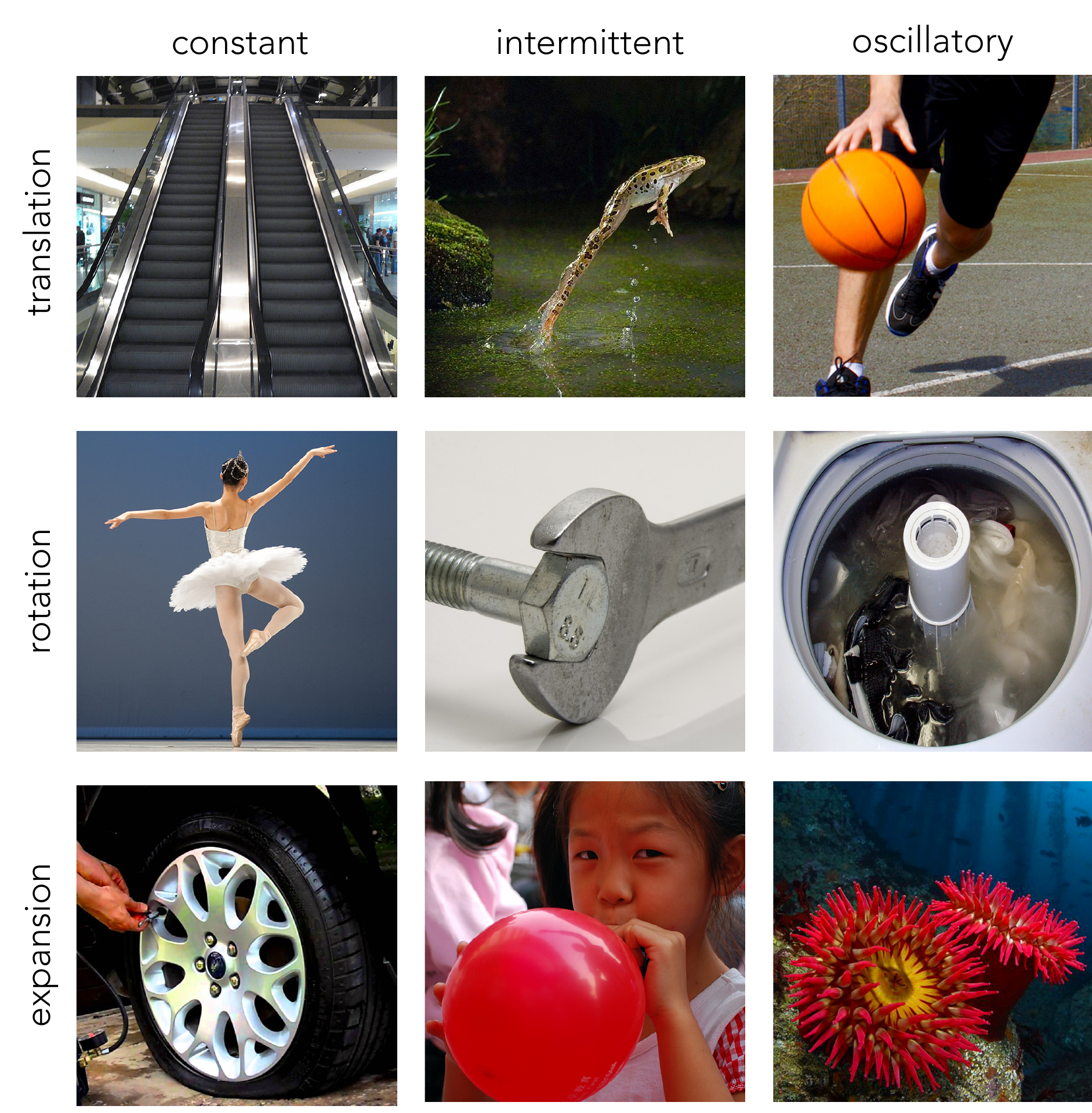}
        \caption{Examples in Real Life}
        \label{fig:3x3_examples}
    \end{subfigure}

    \caption{$3\times 3$ Cartesian table of the \emph{motion type} times the \emph{motion continuity}. These are the basic cases of periodicity in 3D emerging from the motion field decomposition and the temporal dynamics. The examples are: escalator, leaping frog, bouncing ball, pirouette, tightening a bolt, laundry machine, inflating a tire with repetitive texture, inflating a balloon and a breathing anemone.} %Some examples are ubiquitous (\eg constant rotation) while others are hard to find (\eg intermittent expansion).}
    \label{fig:3x3_main_figure}
\end{figure*}

The divergence is a scalar field representing the amount of outward flux from an infinitesimal volume around a given point. Next, the anti-symmetric part $\*R$, referred to as the spin- or rotation matrix, is given by $\*R = \tfrac{1}{2}(\*J - \*J^T)$ with elements:
\begin{align}
  \*R_{ij} = \frac{1}{2}\left(\frac{\partial \mathcal{F}_i}{\partial x_j} - \frac{\partial \mathcal{F}_j}{\partial x_i} \right).
  % \*S^- &=
  % \begin{bmatrix}
  %   0 & \frac{1}{2}\left(\frac{\partial \mathcal{F}_y}{\partial x} - \frac{\partial \mathcal{F}_x}{\partial y}\right) & \frac{1}{2}\left(\frac{\partial \mathcal{F}_z}{\partial x} - \frac{\partial \mathcal{F}_y}{\partial z}\right) \\[6pt]
  %   \frac{1}{2}\left(\frac{\partial \mathcal{F}_x}{\partial y} - \frac{\partial \mathcal{F}_y}{\partial x}\right) & 0 & \frac{1}{2}\left(\frac{\partial \mathcal{F}_z}{\partial y} - \frac{\partial \mathcal{F}_y}{\partial z}\right) \\[6pt]
  %   \frac{1}{2}\left(\frac{\partial \mathcal{F}_x}{\partial z} - \frac{\partial \mathcal{F}_z}{\partial x}\right) & \frac{1}{2}\left(\frac{\partial \mathcal{F}_y}{\partial z} - \frac{\partial \mathcal{F}_z}{\partial y}\right) & 0 \\[6pt]
  % \end{bmatrix}.
\end{align}
From the elements of the spin matrix we can recognize the \emph{curl} of the flow field. More specifically, the curl of the 3D flow field is defined as:
\begin{align}
  \curll = \left[\frac{\partial \mathcal{F}_3}{\partial x_2} - \frac{\partial \mathcal{F}_2}{\partial x_3},\,\, \frac{\partial \mathcal{F}_1}{\partial x_3} - \frac{\partial \mathcal{F}_3}{\partial x_1},\,\, \frac{\partial \mathcal{F}_2}{\partial x_1} - \frac{\partial \mathcal{F}_1}{\partial x_2} \right]^T. \label{eq:curl-definition}
\end{align}
This vector field describes the infinitesimal rotation around a given point. Finally, the last fundamental component is given by the symmetric part $\*E = \tfrac{1}{2}(\*J + \*J^T)$ with elements:
\begin{align}
\*E_{ij} = \frac{1}{2}\left(\frac{\partial \mathcal{F}_i}{\partial x_j} + \frac{\partial \mathcal{F}_j}{\partial x_i} \right).
% \*S^+ =
%   \begin{bmatrix}
%     0 & \frac{1}{2}\left(\frac{\partial \mathcal{F}_y}{\partial x} + \frac{\partial \mathcal{F}_x}{\partial y}\right) & \frac{1}{2}\left(\frac{\partial \mathcal{F}_z}{\partial x} + \frac{\partial \mathcal{F}_y}{\partial z}\right) \\[6pt]
%     \frac{1}{2}\left(\frac{\partial \mathcal{F}_x}{\partial y} + \frac{\partial \mathcal{F}_y}{\partial x}\right) & 0 & \frac{1}{2}\left(\frac{\partial \mathcal{F}_z}{\partial y} + \frac{\partial \mathcal{F}_y}{\partial z}\right) \\[6pt]
%     \frac{1}{2}\left(\frac{\partial \mathcal{F}_x}{\partial z} + \frac{\partial \mathcal{F}_z}{\partial x}\right) & \frac{1}{2}\left(\frac{\partial \mathcal{F}_y}{\partial z} + \frac{\partial \mathcal{F}_z}{\partial y}\right) & 0 \\[6pt]
%   \end{bmatrix}.
\end{align}
This trace-free matrix is known as the deformation tensor and associated with the \emph{shear} of the flow field. In \autoref{fig:quiver-plot-all} we illustrate three motion fields with either pure divergent, rotational or shear flow.

\subsection{Intrinsic Periodic Motion in 3D}
\label{subsec:theory-periodic-motion-in-3d}

\subsubsection{Motion Types}
\label{subsec:motion-types}

For an object moving periodically through the 3D space, the decomposition of the flow field tied to the object is used to characterize the type of motion. A non-rigid object that is expanding or contracting along one or more axes will produce a purely divergent flow field $\divergencee$. Examples include: inflating a balloon or a pulsing anemone. Moreover, a flow field exclusively containing curl $\curl$ emerges with rotational motion such as a spinning wheel or tightening a bolt. Finally, shear is associated with deformation or stress on a surface caused by opposing forces parallel to the cross-section of a body. Shear predominantly plays a role for materials with high-elasticity (\eg fluids) or in the presence of large forces (\eg solid mechanics). Generally, the 3D motion field's shear component is negligible as excessive forces are required to deform the material. For softer materials such as foam, paper or plastics, the shear components can be measurable but this is rare in practice. Based on its rare appearance, we therefore leave the shear for what it is.

In particular, three basic 3D motion types emerge depending on the values of divergence and curl as follows:
\begin{align*}
   \hspace{0.6cm} \text{\emph{translation}:} &\quad \curll(\*x, t) = \*0, \;\;\; \divergencee(\*x, t) = 0 \\
   \,\,\, \text{\emph{rotation}:}    &\quad \curll(\*x, t) \neq \*0, \;\;\; \divergencee(\*x, t) = 0 \\
   \,\,\, \text{\emph{expansion}:}   &\quad \curll(\*x, t) = \*0, \;\;\; \divergencee(\*x, t) \neq 0.
\end{align*}
These motion types are tied to a particular 3D motion field of pure form. In practice there may be a mixture types. As we are aiming to handle realistic video, our method employs a combination of first-order differential motion maps from which the dominant 3D periodicity in the object's motion is determined.

\subsubsection{Motion Continuities}
\label{subsec:motion-continuties}

As periodic motion by its nature contains a temporal component, we here transition to the temporal dynamics of the time-varying motion field. Consecutive measurements of the flow field $\mathcalbf{F}(\*x,t)$ produce a time-varying motion field with particular temporal dynamics. Depending on the type of motion, the motion field needs to satisfy one of the following necessary periodic conditions: %\todo{rewrite $\nabla_t$}:
\begin{align}
   \hspace{0.8cm} \gradd(\*x,t) &= \gradd(\*x + \*\epsilon, t+T)  \nonumber \\
   \curll(\*x,t) &= \curll(\*x + \*\epsilon, t+T) \\
   \divergencee(\*x,t) &= \divergencee(\*x + \*\epsilon, t+T), \nonumber
\end{align}
where $\*\epsilon$ denotes a translation as the object's periodicity may be superposed on translation. For robustness, our method measures both $\mathcalbf{F}(\*x,t)$ and $\gradd(\*x,t)$. From the direction and temporal dynamics of motion, three distinct periodic motion continuities can be distinguished: \emph{constant}, \emph{intermittent} and \emph{oscillating} periodicity. In practice the motion continuity may be a mixture between types. For intermittent and oscillating motion repetitive nature is intrinsically in the temporal dynamics whereas for constant motion to appear repetitively, there will be special conditions on the object's texture or albedo.
% %
% \begin{align*}
%    \hspace{0.8cm}
%    \text{\emph{constant}:}     &\quad \rightarrow \\
%    %& |\mathcalbf{F}(\*x,t)| = |\mathcalbf{F}(\*x,t + \Delta t)| \quad\,\, \forall t \\
%    \text{\emph{intermittent}:} &\quad \rightarrow,\rightarrow \\
%    %& \text{sgn}(\mathcalbf{F}(\*x,t)) = \text{sgn}(\mathcalbf{F}(\*x,t + \Delta t)) \quad \forall t\\
%    \text{\emph{oscillating}:}  &\quad \rightarrow,\leftarrow
% \end{align*}
% %
% In other words, three different periodic motion continuities can be distinguished from the sign changes in motion: constant, intermittent and oscillating periodicity. Again, in practice the motion continuity may be a mixture between types.

\begin{figure*}[t]
  \centering
  \includegraphics[width=1.0\textwidth,clip,trim={3cm 0 3cm 0}]{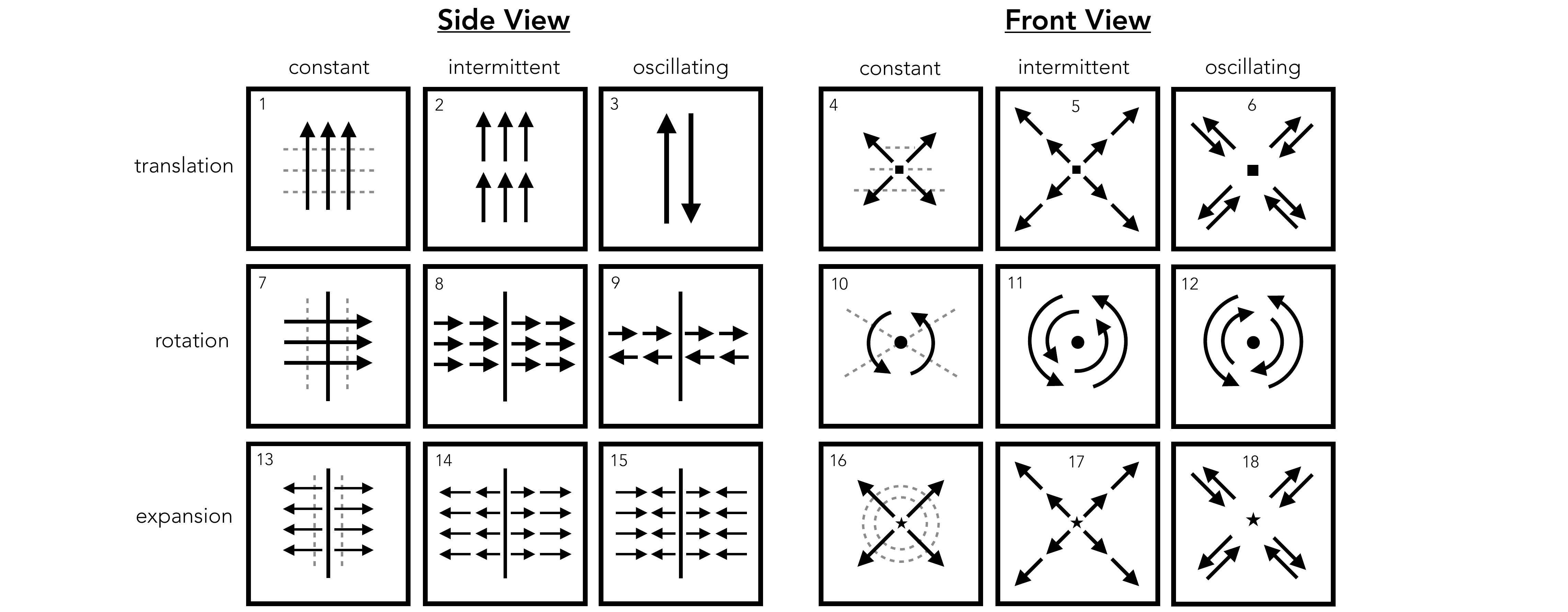}
  \caption{Observed flow: the $18$ fundamental cases for 2D perception of 3D recurrence. The perception follows from the motion pattern ($3\times$), motion continuity ($3\times$) and the viewpoint on the continuous interval between the two extremes: side and front view. $\mathbf{\uparrow}$ denotes flow direction, $\blacksquare$ denotes a vanishing point, $\bullet$ denotes a rotation point, $\bigstar$ denotes expansion point. Dashed grey lines for constant motion indicate the need for texture to perceive recurrence. Pairs $4$-$16$, $5$-$17$ and $6$-$18$ appear similar at first sight but vary in their signal profile.} %\todo{update figure to continuous nature clear.}
  \label{fig:classification-motion-types}
\end{figure*}

\subsubsection{Categorization of Periodic Motion}
\label{subsec:categorization-of-periodic}

The intrinsic periodicity in 3D does not cover all perceived recurrence in an image sequence. For the trivial cases of constant translation and constant expansion in 3D, the perceived recurrence will appear when a repetitive chain of objects (conveyor) or a repetitive appearance (texture on a car tire) on the object is aligned with the motion. In such cases, the recurrence will also be observed in the field of view. For constant rotation, the restriction is that the appearance cannot be constant over the surface, as no motion, let alone recurrent motion would be observed. In the rotational case, any rotational symmetry in appearance will induce a higher order recurrence as a multiplication of the symmetry and the rotational speed. 

For the purpose of periodic motion, nine cases organize in a $3\times3$ Cartesian table of basic \emph{motion type} times \emph{motion continuity}, see \autoref{fig:3x3_cubes}. The corresponding examples of these nine cases are given in \autoref{fig:3x3_examples}. This is the list of fundamental cases, where a mixture of types is permitted. In practice, some cases are ubiquitous, while for others it is hard to find examples at all.

\subsection{Visual Recurrence in 2D}
\label{subsec:recurrence-in-2d}

So far we have considered the intrinsic periodicity in 3D. We reserve the term \emph{recurrent} for the 2D observation of the 3D periodicity. Recurrence in the field of view is defined by:
\begin{align}
  \*{F}(\*x', t) = \*{F}(\*x' + \*\epsilon', t + T)
  \label{eq:flow-2d}
\end{align}
where $\*{F}(\*x', t)$ is the perceived flow in 2D image coordinates $\*x'$. The observed displacement is denoted by $\*\epsilon'$ and $T$ is the temporal period. \emph{The underlying principle is that the same period length $T$ will be observed in both 3D and 2D for all cases of periodicity.} This permits us to measure 3D motion periodicity $T$ from the 2D flow field. Only in some rare cases, the period of motion may change due to a partial or complete occlusion; or the periodic motion disappears entirely due to lack of texture or albedo from a given viewpoint (\eg a constantly rotating textureless disk). These are exceptional cases as the general principle applies that the temporal period is viewpoint invariant.

The camera position relative to the object's motion has a large influence on the perception of the flow field. There are two fundamentally different viewpoints: the \emph{frontal} view and the \emph{side} view:
\begin{align*}
   \text{\emph{frontal view}:} &\quad \text{on the main axis of motion} \\
   \text{\emph{side view}:}    &\quad \text{perpendicular to the main axis of motion}.
\end{align*}
For translation, there is one main axis and two perpendicular axes, which are both identical for our purpose. There is no distinction between the two perpendicular views as their perception is equivalent. Similarly, for rotation, the two perpendicular cases are also indistinguishable. For expansion there are one, two or three axes of expansion, again leaving us with the frontal case and the perpendicular case as the two fundamental cases. Consequently, for all cases considered, a distinction between frontal view and side view is sufficient. As a result, the perceived recurrence is defined on the continuous range between the two extreme viewpoints. Combining the two viewpoint extremes with the nine cases of periodic motion we arrive at the classification of $18$ basic cases as illustrated in \autoref{fig:classification-motion-types}. The two views are the end of a continuous range of viewpoints. Most of the time an actual viewpoint will be somewhere in between the frontal view and the side view. This leaves the flow field asymmetrical or skewed, either in gradient, curl or divergence. As long as $T$ can be measured from the observed signal, the skewed or asymmetric observation will not affect the recurrent nature nor the period of the 3D motion field.

\begin{figure}
\centering
    \begin{subfigure}[b]{0.32\columnwidth}
        \includegraphics[width=\textwidth]{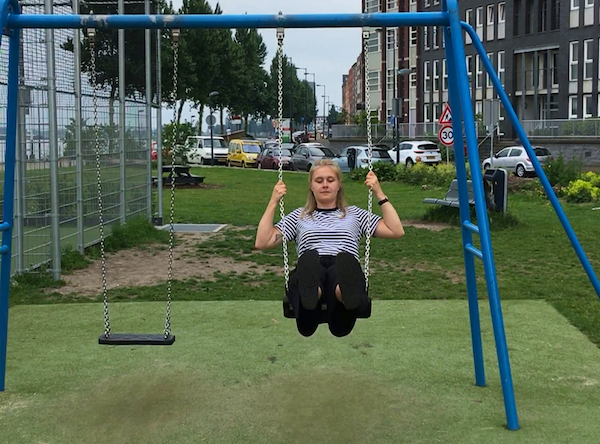}
        \caption{Frontal view}
        \label{fig:gudrun-frontal}
    \end{subfigure}
    %\hspace{0.1em}
    \begin{subfigure}[b]{0.32\columnwidth}
        \includegraphics[width=\textwidth]{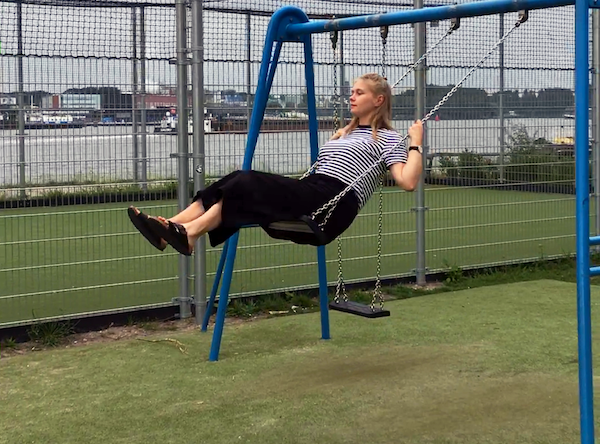}
        \caption{In-between view}
        \label{fig:gudrun-inbetween}
    \end{subfigure}
    %\hspace{0.1em}
    \begin{subfigure}[b]{0.32\columnwidth}
        \includegraphics[width=\textwidth]{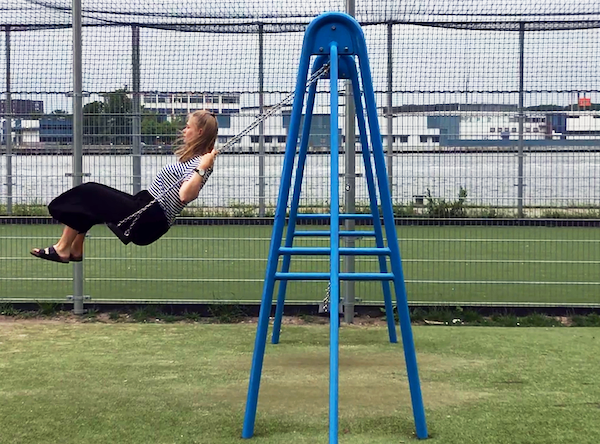}
        \caption{Side view}
        \label{fig:gudrun-side}
    \end{subfigure}
    
    \caption{Example video displaying \emph{girl on a swing} captured from three distinct viewpoints. Moving from one end of the continuous viewpoint spectrum (frontal) to the other (side) results in a dramatic change of motion appearance. The in-between viewpoint leaves the motion measurements either skewed or asymmetrical. In practice, we combine the motion representations to emphasize the one best measurable.}
    \label{fig:gudrun-swing}
\end{figure}

\subsection{Non-Static Repetition}
\label{subsec:moving-viewpoint}

Relative motion between the moving object and the observer adds another dimension of complexity. In particular with recurrent motion (1) the camera may move because the camera is mounted on the moving object itself, or (2) the camera is following the target of interest, or (3) the camera is in motion independent of the motion of the object. For the first two cases, the camera motion reflects the periodic dynamics of the object's motion. The flow field may be outside the object, but otherwise it displays a complementary pattern in the flow field.

\begin{figure*}[t]
  \centering
  \includegraphics[width=\textwidth,trim={0 5.6cm 10.2cm 0},clip]{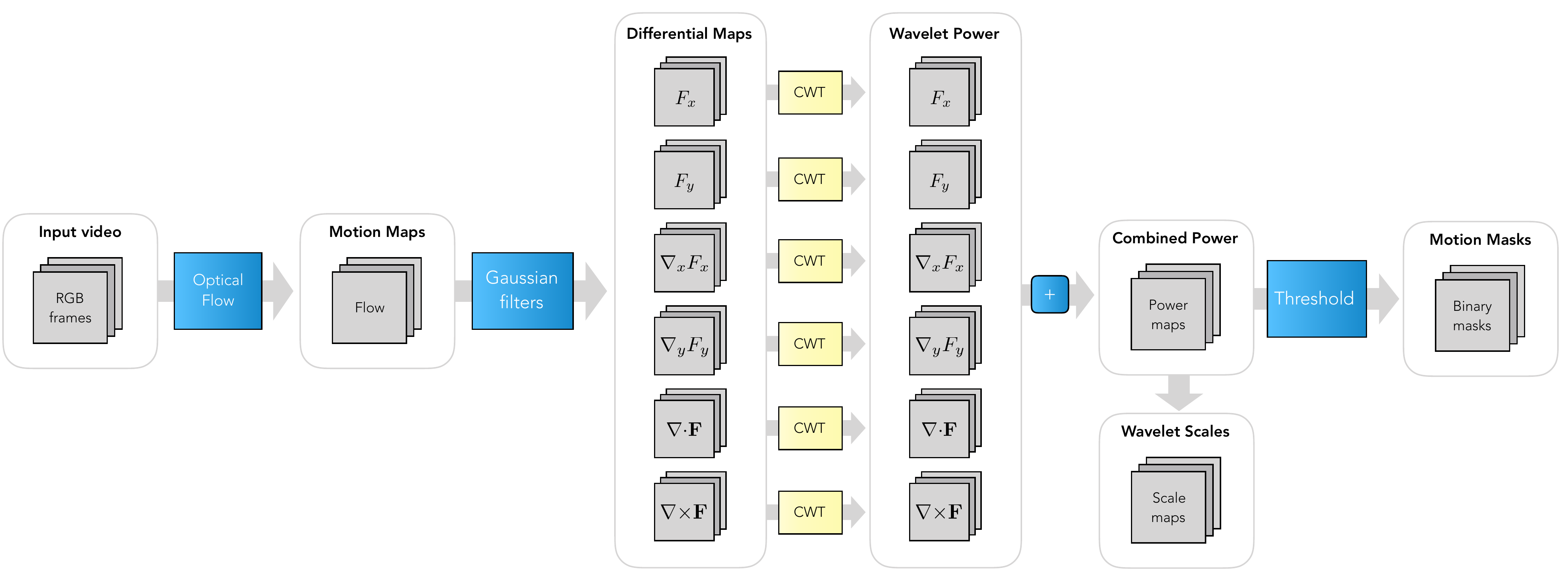}
  \caption{Overview of our method for repetition estimation in video. Given an input video as RGB frames we first estimate the motion between consecutive frames using optical flow. We perform spatial Gaussian filtering to obtain six (differential) motion representations. Next, we apply the continuous wavelet transform (CWT) through temporal convolution over all six representations individually. We combine all power maps by summation to arrive at a single power map for a moment in time. Finally, we spatially segment repetitive motion by mean-thresholding of the power maps. To estimate repetition, we median-pool the wavelet scales over the motion segmentation producing an instantaneous frequency measurement.}
  \label{fig:system_overview}
\end{figure*}

In the first case, the periodically moving camera will produce a global repetitive flow field as opposed to local repetitive flow when the object itself is moving. The third case particularly demands the removal of the camera motion prior to the repetitive motion analysis. In practice, this situation occurs frequently. Therefore, particular attention needs to be paid to camera motion independent of the target’s motion. When the viewpoint changes from frontal to side view due to camera motion, the analysis will be inevitably hard. \autoref{fig:classification-motion-types} illustrates the dramatic changes in the flow field when the camera changes from one extreme viewpoint (side) to the other (frontal), or vice versa. Our method handles such appearance changes by simultaneously using multiple motion representations and summing temporal filter responses.

In addition, even when object motion and camera are both static, for none of the intrinsic motion types (translation, rotation, expansion), a point on the object will be at the same position in the camera field all the time. Under the double static condition, a point will just return to the same point on the camera field. As the intermediate points on the object or background have an arbitrary albedo and radiate an arbitrary luminance, it will not produce a sinusoidal signal in general. This is noteworthy as previous work \citep{cutler2000robust,liu1998finding,pogalin2008visual} implicitly assume such a signal by considering the Fourier transform or variants.

\subsection{Non-Stationary Repetition}
\label{subsec:recurrent-signal-characteristics}

A recurrent signal is said to be stationary when the period length is constant over time. In the initial steps of periodicity analysis, it was assumed the periodic signal was near-stationary. However, decay in frequency or acceleration are common in realistic video. In practice, we have observed that non-stationary is often present, to which we return later with the discussion of our dataset. Therefore, in contrast to \cite{pogalin2008visual} and \cite{levy2015live} we loosen the stationarity assumption, leaving the option of acceleration open. More precisely our method employs the continuous wavelet transform for spectral decomposition of the video.
%!TEX root = ms.tex

\section{Method}
\label{sec:methods}

In this section we present our method for estimating repetition in video. The method takes as input a sequence of RGB frames and outputs a frequency distribution densely computed over space and time. Subsequently, the spectral power distribution, which we obtain from the continuous wavelet transform, is used for repetition counting, motion segmentation or other frequency-based measurements. We target the general case in which moving objects may exhibit non-stationary periodicity or have a non-static appearance due to camera motion or repetition superposed on translation. Our method, summarized in \autoref{fig:system_overview}, comprises motion estimation and two consecutive filtering steps: first we spatially filter the motion fields to arrive at first-order differential geometric motion maps, and then we determine the video's repetitive contents by applying the continuous wavelet transform densely over the motion maps. Task-dependent post-processing steps may give the desired output; here we focus on repetition counting as it enables straightforward evaluation of our method in the presence of non-stationary repetitions.

\subsection{Differential Geometric Motion Maps}
\label{subsec:methods-differential-motion-estimation}

Given a sequence of video frames, we first estimate the motion between pairs of consecutive frames to obtain the motion field $\*F(\*x',t) = (F_x, F_y)$ for all timesteps. Next, the theory implies decomposition of the motion field into the primitive first-order differentials. For a moment in time $t$, we compute the differential motion maps by spatially convolving the flow field with first-order Gaussian derivative filters:
\begin{align}
  %G(\*x ; \sigma) &= \frac{1}{2\pi \sigma^2} \exp\left({-\frac{x^2 + y^2}{2\sigma^2}}\right) \\
  G^x(\*x' ; \sigma) &= -\frac{x}{2\pi \sigma^4} \exp\left({-\frac{x^2 + y^2}{2\sigma^2}}\right) \\
  G^y(\*x' ; \sigma) &= -\frac{y}{2\pi \sigma^4} \exp\left({-\frac{x^2 + y^2}{2\sigma^2}}\right),
\end{align}
where $\sigma$ denotes the spatial scale parameter and image coordinates are given by $\*x' = (x,y)$. Through convolution with Gaussian kernels we obtain the first-order spatial derivatives $\nabla_x F_x, \nabla_y F_x, \nabla_x F_y$ and $\nabla_y F_y$ for a moment in time. Given the spatial partial derivatives of the motion, we compute $\divergence$ and $\curl$ using the 2D equivalents of Eq.~\eqref{eq:divergence-definition} and Eq.~\eqref{eq:curl-definition}. For the 2D case, curl is a single-component vector field perpendicular to the image plane whereas the divergence is a scalar field. To effectively handle all cases of repetitive motion (\autoref{fig:classification-motion-types}), we compute six motion maps for each frame:
\begin{align}
  \big\{\divergence, \curl, \,\,\,\, \nabla_x F_x, \nabla_y F_y, \,\,\,\, F_x, F_y \big\} \label{eq:six-motion-maps}
\end{align}
Periodicity in $\divergence$ or $\curl$ will only occur for the frontal view. For oscillatory or intermittent motion from the side view, $\nabla_x F_x$ and $\nabla_y F_y$ will produce the strongest periodicity while the zeroth-order flow field $F_x$ and $F_y$ will deliver a stronger response for the cases of repetitive periodic appearances at constant motion.

\begin{figure*}

  \vspace{0.1cm}
  \centering

  \begin{subfigure}{0.49\textwidth}
    \centering
    \caption*{\normalsize{\emph{RGB frame}}}
    \includegraphics[width=0.90\textwidth]{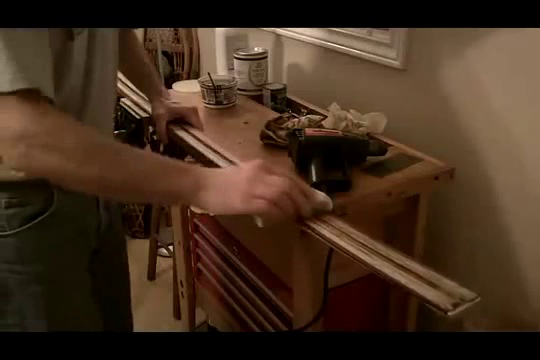}
  \end{subfigure}

  \begin{subfigure}{0.49\textwidth}
    \centering
    \vspace{1em}
    \caption*{\normalsize{$\*F_x$}}
    \vspace{-0.5em}
    \includegraphics[width=0.90\textwidth]{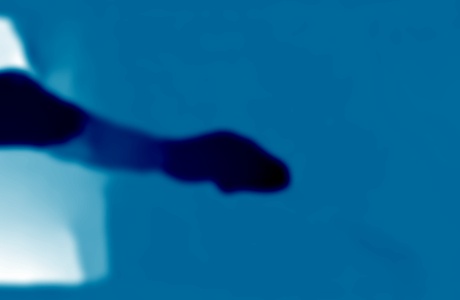}
  \end{subfigure}
  \begin{subfigure}{0.49\textwidth}
    \centering
    \vspace{1em}
    \caption*{\normalsize{$\*F_y$}}
    \vspace{-0.5em}
    \includegraphics[width=0.90\textwidth]{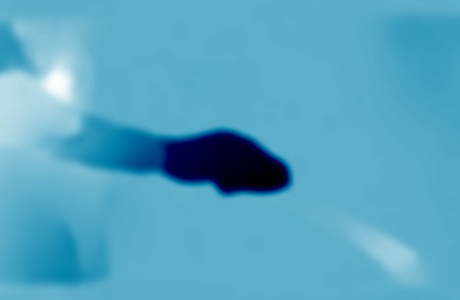}
  \end{subfigure}
  \begin{subfigure}{0.49\textwidth}
    \centering
    \vspace{1em}
    \caption*{\normalsize{$\divergence$}}
    \vspace{-0.5em}
    \includegraphics[width=0.90\textwidth]{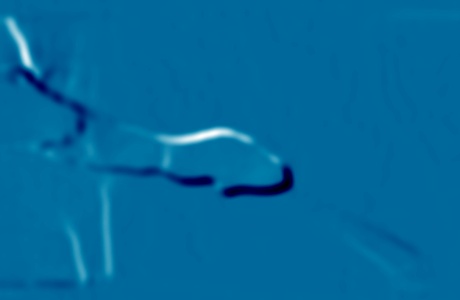}
  \end{subfigure}
  \begin{subfigure}{0.49\textwidth}
    \centering
    \vspace{1em}
    \caption*{\normalsize{$\curl$}}
    \vspace{-0.5em}
    \includegraphics[width=0.90\textwidth]{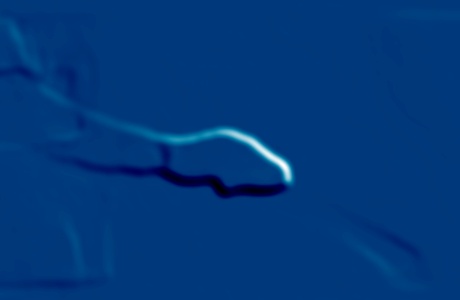}
  \end{subfigure}
  \begin{subfigure}{0.49\textwidth}
    \centering
    \vspace{1em}
    \caption*{\normalsize{\emph{Power Map}}}
    \vspace{-0.5em}
    \includegraphics[width=0.90\textwidth]{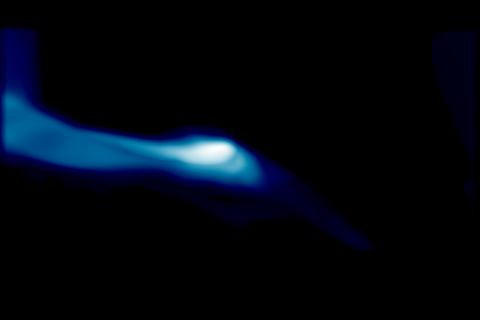}
  \end{subfigure}
  \begin{subfigure}{0.49\textwidth}
    \centering
    \vspace{1em}
    \caption*{\normalsize{\emph{Temporal Scale Map}}}
    \vspace{-0.5em}
    \includegraphics[width=0.90\textwidth]{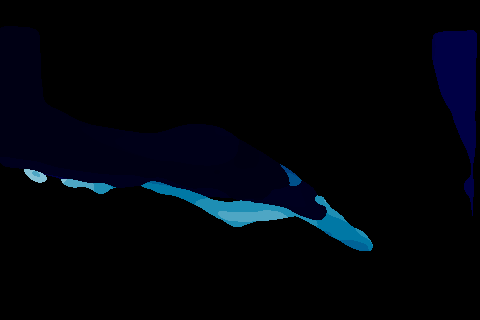}
  \end{subfigure}
  \vspace{1em}
  \caption{Intermediate motion maps for a video displaying a \emph{man brushing wood} from the \datasetname{} dataset displaying a brushing motion. We perform wavelet filtering over six motion maps, due to space constraints only four are shown while $\*\nabla_x\*F_x$ and $\*\nabla_y\*F_y$ are omitted. Notice how the regions with repetitive motion appear in the wavelet power maps. By thresholding the wavelet power map with the mean power we obtain a repetitive motion map. The temporal scale maps indicate spatial regions with motion of low- and high-frequency.}
  \label{fig:motion-maps-example-2}

\end{figure*}

\autoref{fig:motion-maps-example-2} displays an example frame with four of six motion maps (the two are omitted here). The six motion maps represent the video for each moment in time and address the diversity in repetitive motion. In our experiments, we will evaluate the individual and joint representative power associated with the motion maps. \emph{A priori} it is unknown which motion we are dealing with, to which we return later by combining the temporal responses of all motion maps.

\subsection{Dense Temporal Filtering}
\label{subsec:dense-spectral-decomposition}

So far we have only considered spatial filtering to obtain the motion maps for a moment in time. Here we include time and proceed by temporal filtering of the motion maps to estimate the video's repetitive motion. This is where the current method diverges from our previous work. In \citep{runia2018real}, we relied on the same motion maps but performed max-pooling over the foreground motion segmentation obtained separately from \cite{papazoglou2013fast}. The max-pooled values over time construct a one-dimensional signal acting as a surrogate for the dynamics in a particular motion map. Spectral decomposition for each of the signals led to six (possibly contrasting) time-frequency estimates. To select the most discriminative representation, we employed a self-quality assessment based on the spectral power in the signals. 

We found two problems with this approach: (1) the decoupled motion segmentation may not be optimal for estimating repetitive motion dynamics, and (2) max-pooling over the foreground motion mask discards most information and is unable to deal with multiple moving parts. We here address these problems by dense temporal filtering over all locations in the motion map instead of operating on the max-pooled signals. Spatially dense estimation of the local spectral power enables us to localize regions likely containing repetitive motion. The temporal filtering can be implemented in several ways, for example, as Fourier transform through temporal convolution. To handle non-stationary video dynamics, we perform the continuous wavelet transform by convolution to obtain a time-varying spectral decomposition.

%%%%%%%%%%%%%%%%%%%%%%%%%%%%%%%%%%%%%%%%%%%%%%%%%%%%%%%%%%%%%%%%%%%%%%%%%%%%%%%%
%%%%%%%%%%%%%%%%%%%%%%%%%%%%%%%%%%%%%%%%%%%%%%%%%%%%%%%%%%%%%%%%%%%%%%%%%%%%%%%%

\subsection{Continuous Wavelet Transform}
\label{subsec:methods-continuous-wavelet-transform}

% \todo{move?} Specifically, we compute the continuous wavelet transform through temporal convolution over all locations in the motion map resulting in a spatially dense time-frequency representation. As our aim is to handle non-stationary video, the wavelet transform is preferred over Fourier analysis for non-stationary video. Moreover, the well-known Short-Time Fourier Transform performs the Fourier transform in a sliding-window fashion but falls short in effectively dividing the time-frequency space to perform a true multi-scale analysis. We proceed with a brief discussion on the continuous wavelet transform that we employ for spectral decomposition.

\begin{figure}
  \centering
  \includegraphics[width=\columnwidth]{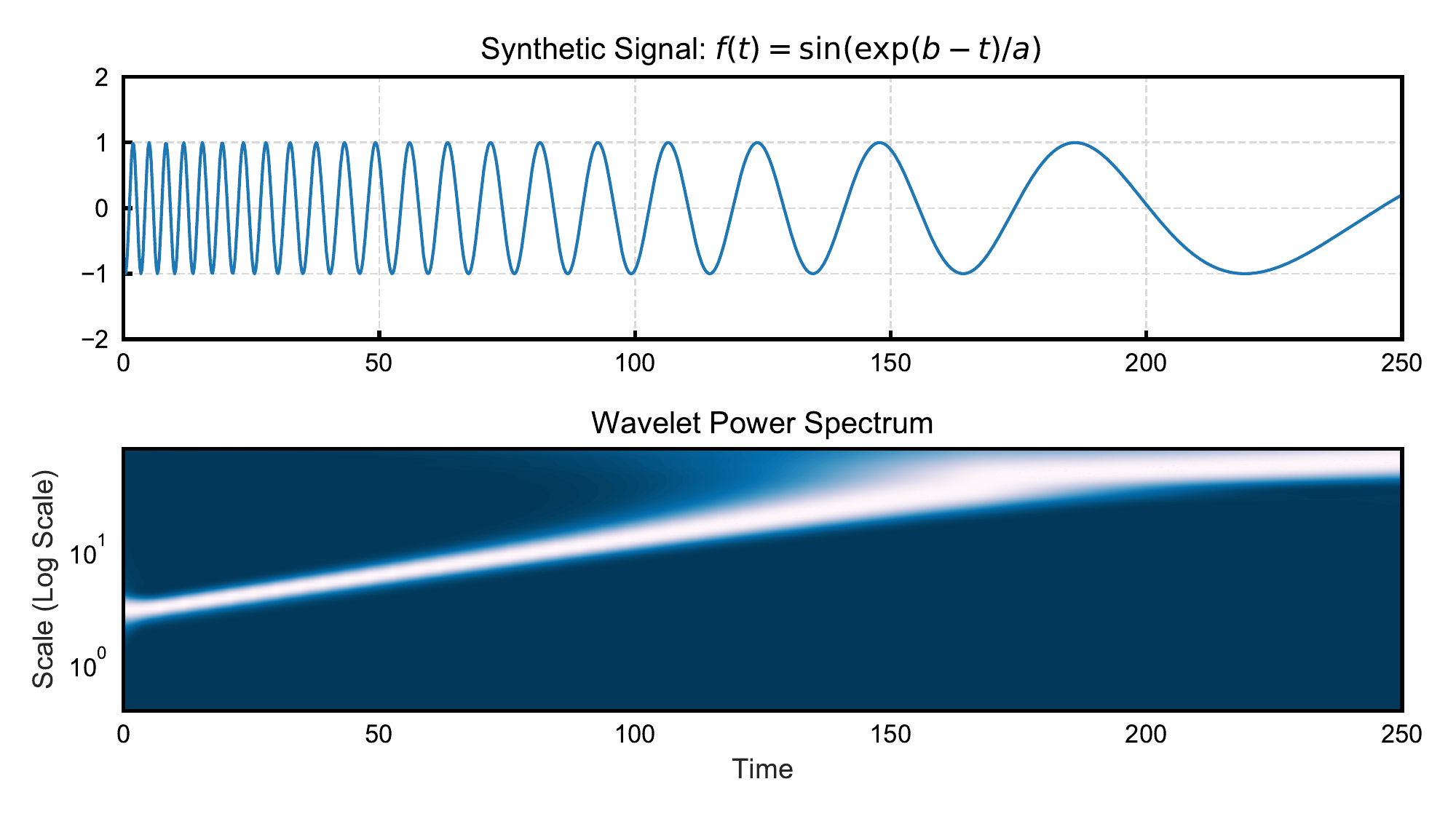}
  \vspace{-2em}
  \caption{Exponential chirp signal and the corresponding scalogram obtained from the continuous wavelet transform. Note increasing scale (period) in the scalogram as the signal's frequency decreases.}
  \label{fig:exponential-chirp-signal}
\end{figure}

% Classical signal processing methods such as the Fourier transform are not suitable for analyzing non-stationary signals. As real-world video is rarely perfectly periodic these methods and inapt. The well-known Short-Time Fourier Transform performs the Fourier transform in a sliding-window fashion but falls short in effectively dividing the time-frequency space in order to perform true multi-scale analysis. As a remedy, we propose to use the continuous wavelet transform for handling non-stationary video with changing frequency or transient (non-stationary) phenomena.

\noindent \,\, Given a discrete signal $h_n$ for timesteps \mbox{$n = 1, \ldots, N-1$} sampled at equally spaced intervals $\delta t$. Let $\psi_0(\eta)$ be some admissible wavelet function, depending on the non-dimensional time parameter $\eta$. The continuous wavelet transform \citep{grossmann1984decomposition} is defined as the convolution of $h_n$ with a ``daughter'' wavelet generated by scaling and translating the wavelet function $\psi_0(\eta)$:
\begin{equation}
\label{eq:continuous-wavelet-transform}
W_n(s) = \sum_{n'=0}^{N-1} h_{n'} \psi^*\left[\frac{(n'-n)\delta t}{s} \right],
\end{equation}
where the asterisk represents the complex conjugate. By varying time parameter $n$ and the scale parameter $s$, the wavelet transform generates a time-scale representation describing how the amplitude of the signal changes with time and scale. We use the Morlet wavelet, a complex exponential carrier modulated by a Gaussian envelope:
%While formally a time-scale representation, it can also be considered a time-frequency representation since the wavelet scale is directly related to the Fourier frequency.
%
\begin{equation}
\label{eq:morlet-wavelet}
\psi_0(\eta) = \pi^{-1/4} e^{i \omega_0 \eta} e^{\eta^2 / 2}.
\end{equation}
In all our experiments we set $\omega_0 = 6$ as it provides a good balance between time and frequency localization. Since the Morlet wavelet is complex, the wavelet transform $W_n(s)$ is also complex. Therefore, it is useful to define the wavelet power spectrum or \emph{scalogram} as $|W_n(s)|^2$ representing the time-frequency localized energy. \autoref{fig:exponential-chirp-signal} gives a non-stationary signal example and plots its wavelet power. It is clear that the scalogram is effective in revealing the signal's non-stationary repetitive dynamics.

The resolution of the scalogram $|W_n(s)|^2$ is defined by the distribution of scale parameter $s$. In practice, we use a discrete scale set that is logarithmically distributed:
\begin{align}
  s_j &= s_02^{j \delta j}, \quad j = 0,1,\ldots,J \label{eq:scale-distr1} \\
  J   &= \delta j^{-1} \log_2 \left( N \delta t / s_0 \right) \label{eq:scale-distr2}.
\end{align}
The smallest measurable scale $s_0$ and the number of scales $J$ determines the range of the detectable frequencies. The smallest scale should be chosen such that the Fourier period of the wavelet is approximately $2\delta t$.

For a moment in time, the scalogram's maximum power will give the wavelet scale $s$ producing the strongest filter response. Often the temporal frequency associated with the scale $s$ will be a more convenient measurement. Therefore, the wavelet scale can be converted to a temporal frequency. For a Morlet wavelet, the relationship between scale and wavelength is given by \citep{torrence1998practical}:
\begin{align}
  \lambda = \frac{4\pi}{\omega_0 + \sqrt{2+\omega^2}}, \label{eq:wavelet-scale-conversion}
\end{align}
where $\omega_0$ corresponds to the non-dimensional frequency. For $\omega_0 = 6$ corresponds to $\lambda = 1.03s$ for the Morlet wavelet, thus having the attractive property of wavelet scale being almost identical to the wavelength. We use \eqref{eq:wavelet-scale-conversion} to obtain the frequency estimate for each time $t$ and location $\*x'$.

\begin{figure*}

  \centering

   \begin{subfigure}{0.16\textwidth}
    \centering
    \caption*{RGB}
    \vspace{-0.5em}
    \includegraphics[width=\textwidth]{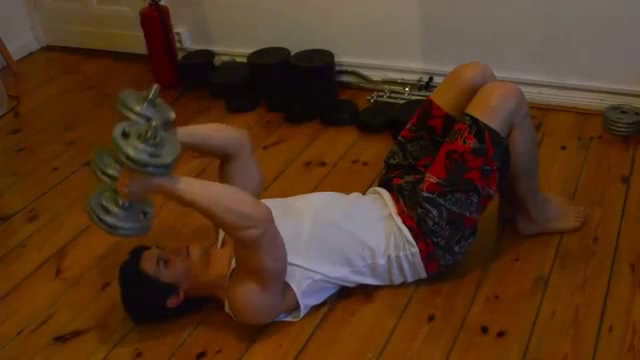}
  \end{subfigure}
  \hspace{0.01em}
  \begin{subfigure}{0.16\textwidth}
    \centering
    \caption*{$F_x$}
    \vspace{-0.5em}
    \includegraphics[width=\textwidth]{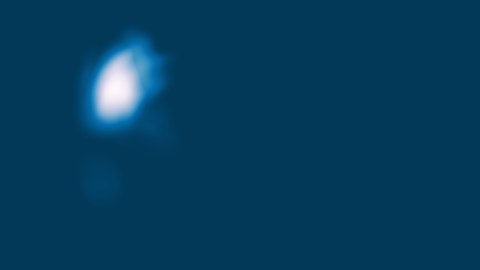}
  \end{subfigure}
  \hspace{0.01em}
  \begin{subfigure}{0.16\textwidth}
    \centering
    \caption*{$F_y$}
    \vspace{-0.5em}
    \includegraphics[width=\textwidth]{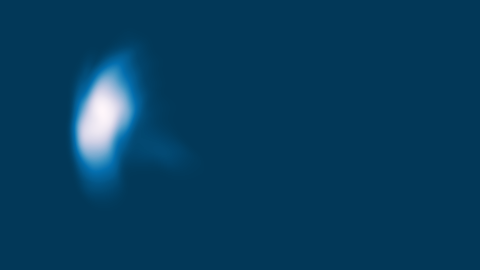}
  \end{subfigure}
  \hspace{0.01em}
  \begin{subfigure}{0.16\textwidth}
    \centering
    \caption*{$\divergence$}
    \vspace{-0.5em}
    \includegraphics[width=\textwidth]{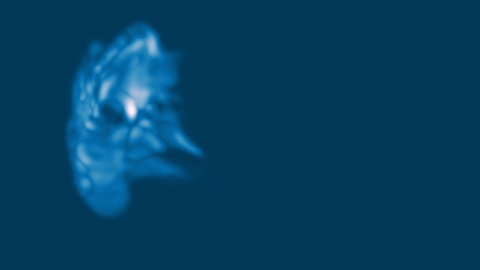}
  \end{subfigure}
  \hspace{0.01em}
  \begin{subfigure}{0.16\textwidth}
    \centering
    \caption*{$\curl$}
    \vspace{-0.5em}
    \includegraphics[width=\textwidth]{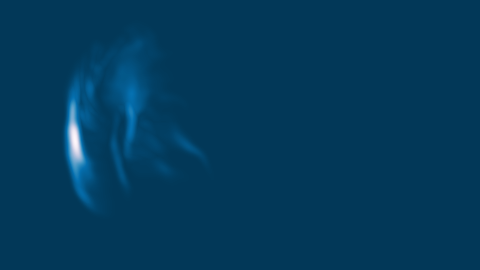}
  \end{subfigure}
  \hspace{0.01em}
  \begin{subfigure}{0.16\textwidth}
    \centering
    \caption*{Total}
    \vspace{-0.5em}
    \includegraphics[width=\textwidth]{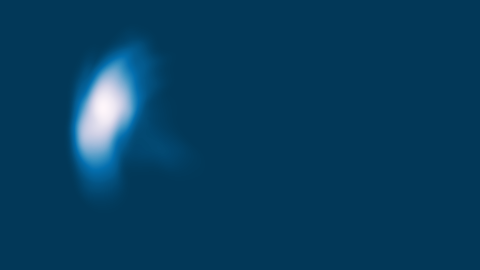}
  \end{subfigure}

  \caption{Video displaying \emph{a man lifting weights} from our video dataset and its corresponding wavelet power maps for individual representations (we omit $\nabla_x F_x$ and $\nabla_y F_y$) . On the right, the total wavelet power obtained through the summation of all six responses. We normalize the power maps for displaying purpose. The vertical flow and curl produce the power maps with the largest norm for this moment in time. Summation of the individual power map combines the responses by emphasizing on the strongest repetitive motion appearance.}
  \vspace{-1em}
  \label{fig:power-maps-individual}

\end{figure*}

\subsection{Combining Spectral Power Maps}
\label{subsec:combining-wavelet-power}

We compute the time-localized frequency estimates by temporal convolution densely over the six individual motion representations. For each representation this produces a time-varying maximum \emph{power map} and \emph{scale map}. The power map contains the spatial distribution of maximum wavelet power over all temporal scales; the scale map holds the temporal scales corresponding to the wavelets with maximum power. What remains is combining the wavelet responses from all motion representations. 

Rather than selecting the single most discriminative representation \citep{runia2018real}, we combine the spectral power maps by summation on a per-frame basis. To illustrate this, we visualize four (out of six) individual power maps and their combined response in \autoref{fig:power-maps-individual}. Summation of the spectral power maps has a number of attractive properties. Most importantly, the motion maps with the strongest repetitive appearance will contribute most to the final power map whereas weakly-periodic motion maps will have a negligible contribution. This effectively serves as a dynamic selection of the most discriminative motion representation. Moreover, as the spectral power is time-localized, the relative contribution per motion representation will be evolving over time. This is appealing because motion appearance can be non-static in realistic video due to camera motion or gradual change in motion type. 

\subsection{Spatial Segmentation}
\label{subsec:methods-spatial-localization}

The combined wavelet power map gives a time-varying spatial distribution of spectral power over all motion representations, whereas the corresponding effective scale map relates to the temporal scale with maximum spectral power. We propose to use the spatial distribution of spectral power for segmentation of the regions with strongest repetitive appearance. Subsequently, we use the scale map to infer the dominant temporal scale (related to the motion frequency) over the localized region.

The spatial segmentation of repetitive motion is performed in a straightforward manner. For a moment in time, we simply mean-threshold the combined wavelet power map to obtain a binary segmentation mask associated with regions containing significant spectral power. More precisely, the wavelet-based motion segmentation will attend to regions in which the maximum spectral power over all temporal scales is significant. \autoref{fig:motion-maps-example-2} (bottom row) illustrates this by displaying the combined power map and corresponding scale map. In general, performing motion segmentation directly from the spatial distribution of spectral power is appealing as it couples the localization and subsequent frequency measurements. Our experiments will verify this claim and compare them with specialized motion segmentation methods. We would like to mention that our segmentation method leaves the door open for multiple repetitively moving objects whereas most state-of-the-art segmentation methods assume a single dominant foreground motion \citep{tokmakov2017}.

% From \autoref{fig:classification-motion-types} it is clear that the practice of measuring recurrence is hard when the viewpoint is moving with the camera or even when it is a mixture between the extreme cases. When the viewpoint is in between purely frontal view
% and side view, the signal will be distorted to a skew form while the divergence or curl-signal of the frontal view will vanish in side view.

% \todo{comment on dense filtering over motion maps again.} This demands more compute resources, which we handle with a new implementation that performs both spatial and temporal filtering on the GPU.

% \begin{figure}
%   \centering
%   \includegraphics[width=\columnwidth,trim={0 7cm 4cm 0},clip]{motion-maps}
%   \caption{An example video frame with corresponding wavelet power map and effective temporal scale map for the $F_y$ representation. Thresholding the power map produces a binary mask identifying regions with strong repetitive motion. To determine the instantaneous frequency of the foreground motion, we median-pool the effective wavelet scales over motion mask.}
%   \label{fig:intermediate-motion-maps}
% \end{figure}

\subsection{Repetition Counting}
\label{subsec:methods-repetition-counting}

%The proposed framework enables localization and identification of repetitive motion in video which can serve a purpose in many vision tasks. For our experiments, we focus on the task of repetition counting. This enables straightforward quantitative evaluation of our method as the number of cycles is well-defined even in the presence of non-stationary motion. We note that by thresholding the power maps we implicitly focus on counting the most  salient repetitive action. Equipped with the motion segmentation mask and effective scale mask, we here discuss our counting mechanism.

To obtain an instantaneous frequency estimate of the salient motion, we median-pool the temporal wavelet scales over the segmentation mask. Median-pooling is preferred over mean-pooling as it relatively robust to outliers and will produce a better estimate of the dominant frequency. The corresponding temporal wavelet scale is then converted to an instantaneous frequency using Eq.~\ref{eq:wavelet-scale-conversion}. For a moment in time, this will deliver a frequency estimate for the salient repetitive motion. Counting the number of repetitions follows temporal integration of the consecutive frequency measurements with the temporal sampling spacing inferred from the video's frame rate.

We emphasize our method's ability to count the number of cycles in non-stationary video. For a stationary periodic signal, the median-pooled temporal scales will be constant over time, while non-stationarity motion produces time-varying frequency estimates. Although the videos considered in our experiments are temporally segmented, the time-localized wavelet responses could also be used for temporal localization of repetitive actions. Moreover, although the current approach performs median-pooling over the motion segmentation mask, the spatial distribution of wavelet power also enables the identification of multiple periodically moving parts. 

%!TEX root = ms.tex

%%%%%%%%%%%%%%%%%%%%%%%%%%%%%%%%%%%%%%%%%%%%%%%%%%%%%%%%%%%%%%%%%%%%%%%%%%%%%%%%

\begin{figure*}

  \begin{subfigure}{\textwidth}
    \centering
    \includegraphics[width=\textwidth]{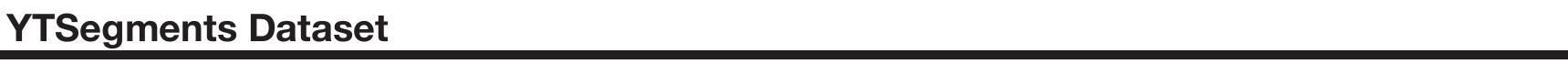}
  \end{subfigure}
  \\[0.3em]
  \begin{subfigure}{\textwidth}
    \centering
    \includegraphics[width=\textwidth,trim={0 9cm 0 0 },clip]{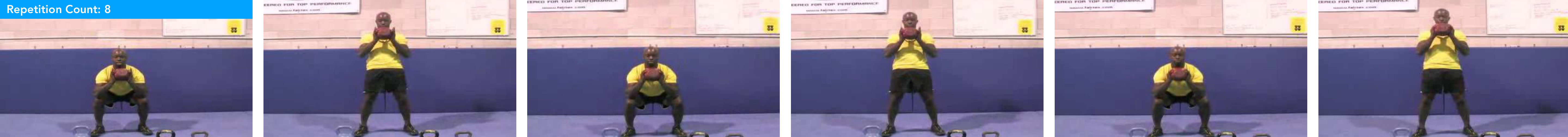}
  \end{subfigure}
  \\[0.3em]
  \begin{subfigure}{\textwidth}
    \centering
    \includegraphics[width=\textwidth,trim={0 9cm 0 0 },clip]{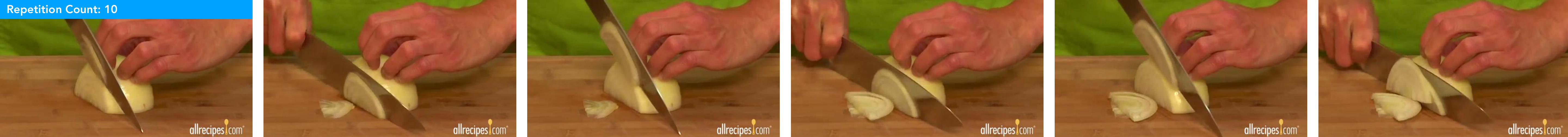}
  \end{subfigure}
  \\[0.3em]
  \begin{subfigure}{\textwidth}
    \centering
    \includegraphics[width=\textwidth,trim={0 9cm 0 0 },clip]{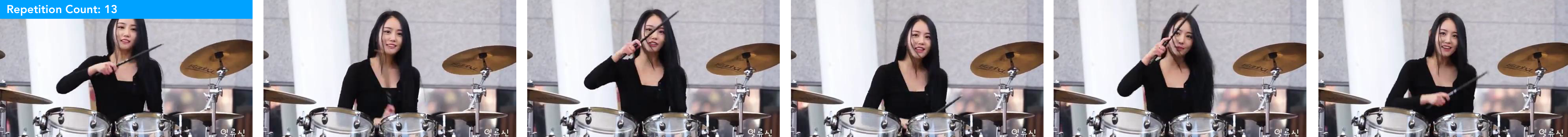}
  \end{subfigure}
  \\[0.3em]
  \begin{subfigure}{\textwidth}
    \centering
    \includegraphics[width=\textwidth,trim={0 9cm 0 0 },clip]{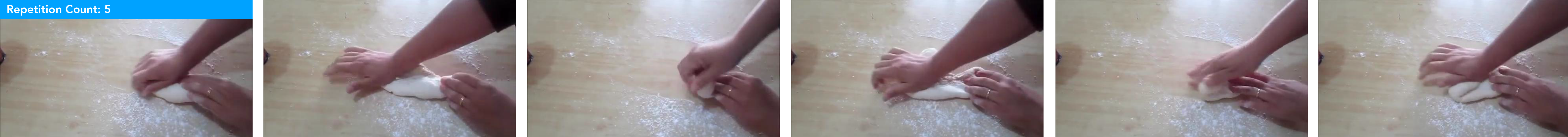}
  \end{subfigure}
  \\[2em]
  \begin{subfigure}{\textwidth}
    \centering
    \includegraphics[width=\textwidth]{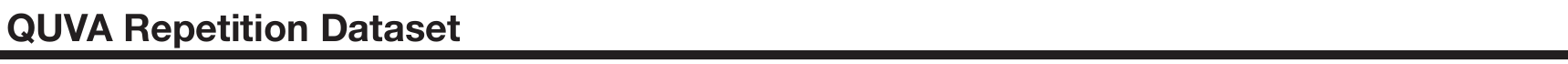}
  \end{subfigure}
  \\[0.3em]
  \begin{subfigure}{\textwidth}
    \centering
    \includegraphics[width=\textwidth,trim={0 9cm 0 0 },clip]{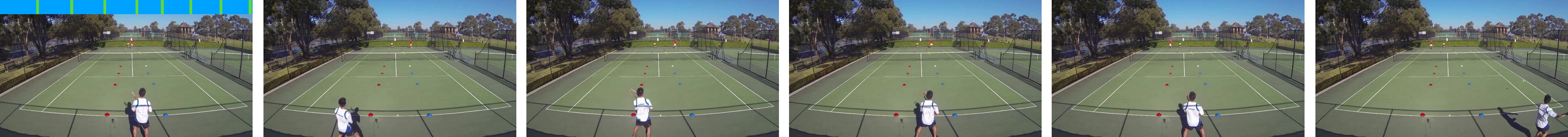}
  \end{subfigure}
  \\[0.5em]
  \begin{subfigure}{\textwidth}
    \centering
    \includegraphics[width=\textwidth,trim={0 9cm 0 0 },clip]{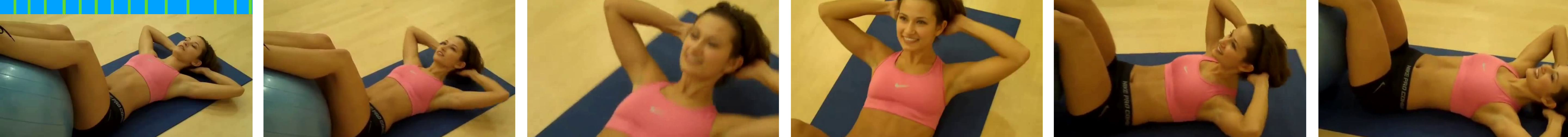}
  \end{subfigure}
  \\[0.5em]
  \begin{subfigure}{\textwidth}
    \centering
    \includegraphics[width=\textwidth,trim={0 9cm 0 0 },clip]{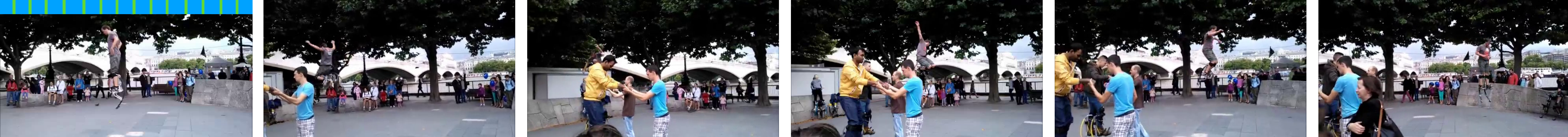}
  \end{subfigure}
  \\[0.5em]
  \begin{subfigure}{\textwidth}
    \centering
    \includegraphics[width=\textwidth,trim={0 9cm 0 0 },clip]{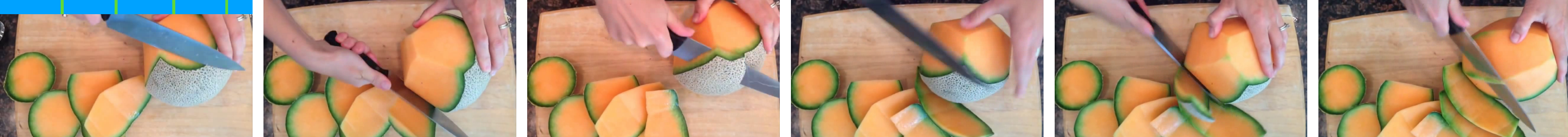}
  \end{subfigure}
  \\[0.5em]
  \begin{subfigure}{\textwidth}
    \centering
    \includegraphics[width=\textwidth,trim={0 9cm 0 0 },clip]{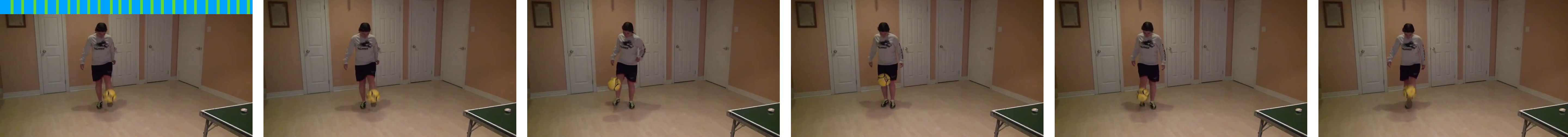}
  \end{subfigure}
  \\[0.5em]
  \begin{subfigure}{\textwidth}
    \centering
    \includegraphics[width=\textwidth,trim={0 9cm 0 0 },clip]{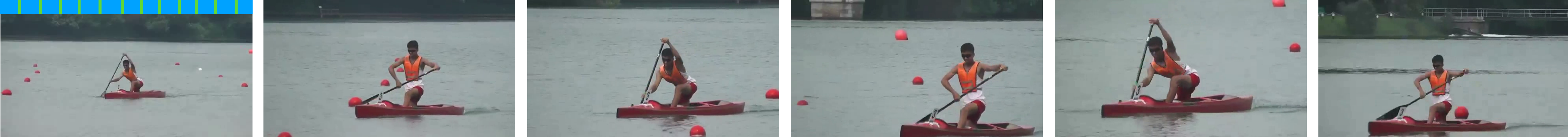}
  \end{subfigure}
  \\[0.5em]
  \begin{subfigure}{\textwidth}
    \centering
    \includegraphics[width=\textwidth,trim={0 9cm 0 0 },clip]{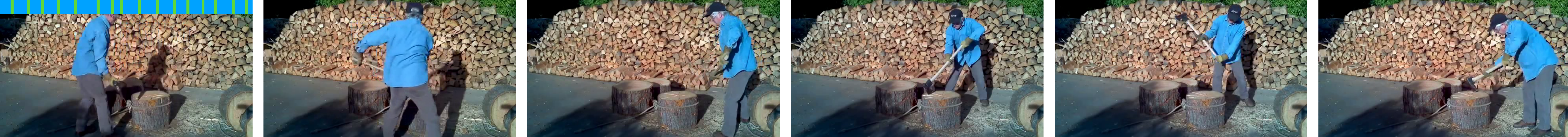}
  \end{subfigure}
  \\[0.5em]
  \begin{subfigure}{\textwidth}
    \centering
    \includegraphics[width=\textwidth,trim={0 9cm 0 0 },clip]{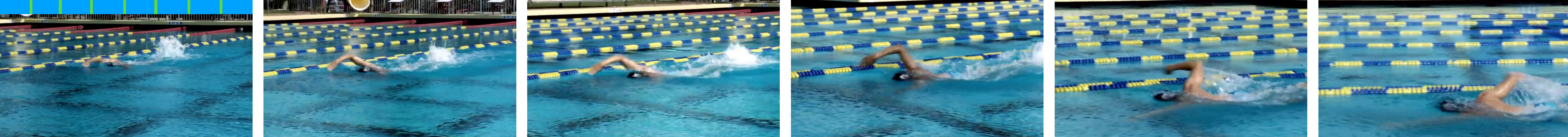}
  \end{subfigure}

  \caption{Four examples from the \ytsegments{} dataset \citep{levy2015live} and eight examples from our \datasetname{} dataset. The \ytsegments{} dataset as released by the authors features a final repetition count annotation (indicated); Our dataset is additionally annotated with individual cycle bounds suitable for determining the level of non-stationarity. The blue timeline in the first frame displays the individual cycle annotations for the given video. The final count is determined by summing the number of individual cycles. Note variation in cycle length and the increased difficulty of our dataset due to camera motion, occlusions and background clutter.}
  \label{fig:dataset-examples}
\end{figure*}

\section{Experiments}
\label{sec:experiments}

We perform experiments to show the effectiveness of our method on the task of counting repetitions in video. Prior to evaluating our full method, we demonstrate the strength of the continuous wavelet transform for estimating repetition in non-stationary signals, show the need for diversified motion maps to deal with the wide variety in motion appearance, and investigate our method's ability to handle dynamic viewpoints. Before discussing the actual experiments, we introduce the video datasets for testing, give implementation details and specify our counting evaluation metrics.

%%%%%%%%%%%%%%%%%%%%%%%%%%%%%%%%%%%%%%%%%%%%%%%%%%%%%%%%%%%%%%%%%%%%%%%%%%%%%%%%
%%%%%%%%%%%%%%%%%%%%%%%%%%%%%%%%%%%%%%%%%%%%%%%%%%%%%%%%%%%%%%%%%%%%%%%%%%%%%%%%
%%%%%%%%%%%%%%%%%%%%%%%%%%%%%%%%%%%%%%%%%%%%%%%%%%%%%%%%%%%%%%%%%%%%%%%%%%%%%%%%

\begin{figure*}
\centering
\begin{minipage}{.48\textwidth}
  \centering

  \captionof{table}{Dataset statistics of \ytsegments{} \citep{levy2015live} and \datasetname{}. The cycle length variation is defined as the average value of the absolute difference between the minimum and maximum cycle length divided by the average cycle length. To determine this, we annotate all individual cycle bounds for both datasets. The last two rows are also obtained by manual annotation.\label{tab:dataset-statistics}}
  % Note that our dataset is more realistic and challenging in terms of cycle length variability, camera motion and motion complexity

  \begin{tabular}{lrr}
    \toprule
                          & \ytsegmentsbold{} & \datasetnamebold{} \\
    \midrule
    Number of Videos      & $100$          & $100$             \\
    Duration Min/Max (s)  & $2.1$/$68.9$ & $2.5$/$64.2$ \\
    Duration Avg. (s)     & $14.9 \pm 9.8$ & $17.6 \pm 13.3$ \\
    Count Avg.$\pm$ Std.  & $10.8 \pm 6.5$ & $12.5 \pm 10.4$   \\
    Count Min/Max         & $4$/$51$       & $4$/$63$               \\
    %Cycle Length Min/Max (frames)  & ? & $8$/$231$ \\
    %Cycle Length Min/Max (s)       & ? & $0.2$/$7.7$ \\
    Cycle Length Variation & $0.22$  & $0.36$            \\
    Camera Motion          & $21$ & $53$ \\
    Superposed Translation & $7$ & $27$ \\
    \bottomrule
    \end{tabular}

\end{minipage}
\quad\quad
\begin{minipage}{.48\textwidth}
  \centering
  \includegraphics[width=\columnwidth]{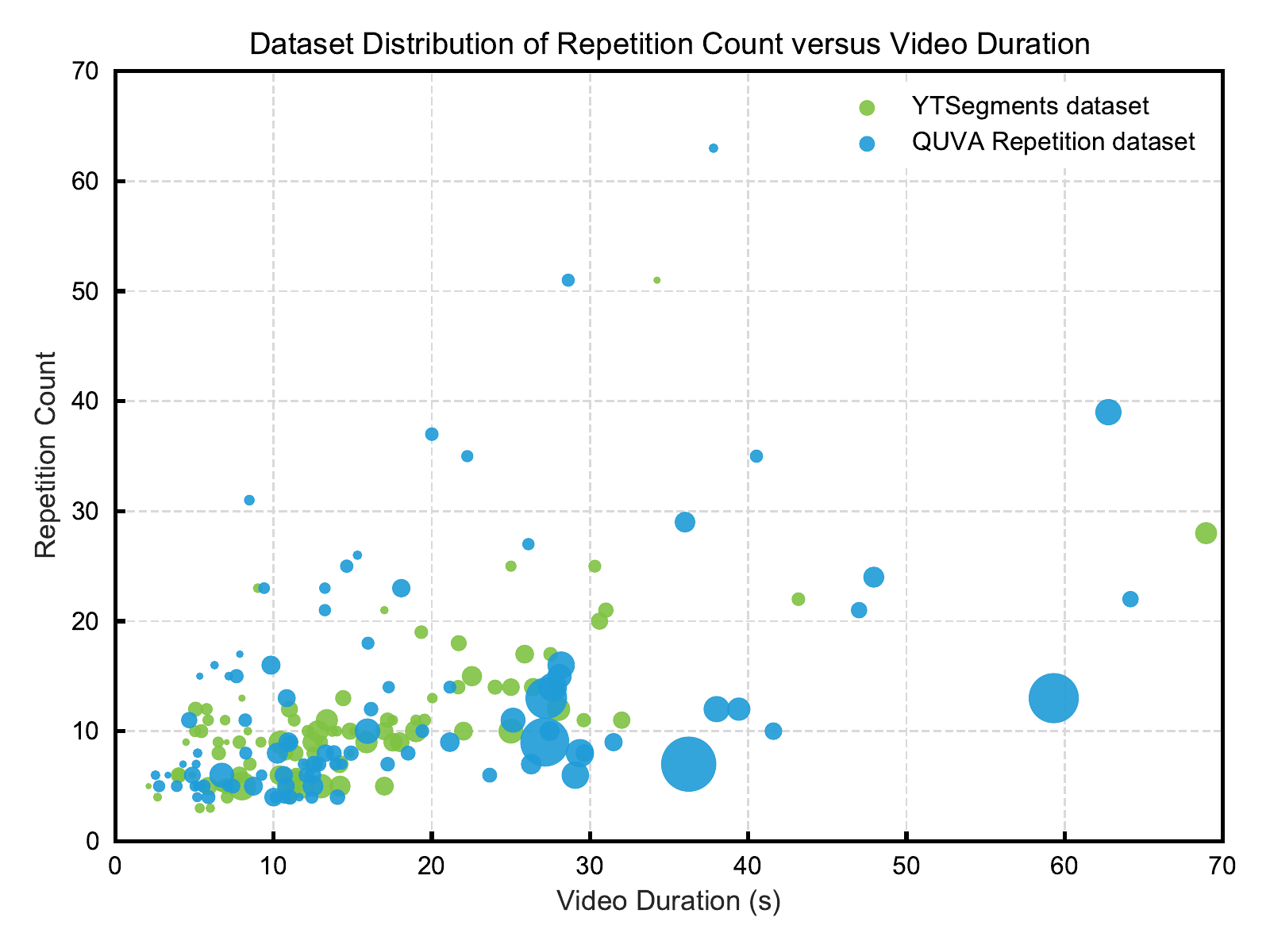}
  \vspace{-2em}
  \caption{Distribution of repetition count versus video duration for the \ytsegments{} and \datasetname{} dataset. The radius of each datapoint is proportional to the cycle length variation of the video. Note the increased variability in non-stationarity and repetition count of our dataset in comparison to \ytsegments{}.
  \label{fig:dataset-distribution}}
\end{minipage}
\end{figure*}

\subsection{Datasets and Evaluation}
\label{subsec:datasets}

The main experiments consider two video datasets: the existing \ytsegments{} and our new \datasetname{} dataset; both collected for the purpose of evaluating repetition estimation in video. Additionally, we perform a controlled experiment on viewpoint estimation with synthetic video that we generated through 3D modeling in Blender. \\ 

\noindent \textbf{YTSegments Dataset.} For the purpose of evaluating repetition counting in video, \cite{levy2015live} introduced a new video benchmark. The $100$ videos downloaded from YouTube are purely for evaluation purpose as training the network is performed with synthesized videos. A wide range of actions appears in the videos: several sports, cooking and animal movement. Each video is temporally segmented such that only the repetitive action is covered. The clips are annotated with a total repetition count. While the dataset serves as a good initial benchmark for repetition estimation, it is limited in terms of cycle length variation (non-stationarity), motion appearances and camera motion. As our goal is to evaluate our method on more realistic video, we introduce a new video dataset that is more challenging in terms of non-stationarity, motion appearance, camera motion and background clutter. \\

% Paragraph: introduction of the dataset, maybe needs a little more motivation
\noindent \textbf{QUVA Repetition Dataset.} In \cite{runia2018real} we introduced a more realistic video benchmark for repetition estimation. The \datasetname{} consists of $100$ videos displaying a wide variety of repetitive video dynamics, including various kinds of sport, music-making, cooking, grooming, construction and animal behavior. The videos are collected from YouTube with emphasis on creating a diverse collection of videos suitable for evaluating our method's ability to deal with non-stationary motion, camera motion and significant evolution of motion appearance over the course of a video.

After video collection, we adopt a multi-stage annotation process to obtain the final dataset. First, we asked two human annotators to label the temporal bounds of each interval containing at least four unambiguous repetitions. We found high inter-agreement between the annotators and keep the $100$ intervals with the highest overlap to increase clarity. Final video clips are obtained by temporal clipping of the intersection of the two intervals. As a result, some motion cycles may be partial either at the beginning or end of the video. In the last round of annotation, we ask the annotators to mark all individual cycle bounds in the video clips (also producing the final repetition count). We also mark the individual cycle bounds for the videos of the \ytsegments{} dataset to compare the inter-cycle length variability representing the level of non-stationarity.

% hack for \ref as the table is in a minipage
The characteristics for both datasets are reported in \hyperref[tab:dataset-statistics]{Table~1}. It is apparent that our videos have more variability in cycle length, motion appearance, camera motion and background clutter. The increased difficulty in both appearance and temporal dynamics give a more realistic benchmark for repetition estimation in the wild. \autoref{fig:dataset-examples} displays a number of examples from both datasets. The project page\footnote{\url{http://tomrunia.github.io/projects/repetition/}} contains the dataset download link and several video previews. \\

\noindent \textbf{Evaluation Metrics.} Given a set of $N$ videos, we evaluate the performance between ground truth count $c_i$ and the count prediction $\widehat{c}_i$ for all videos $i \in \{1,\ldots,N\}$. We report the mean absolute error following prior work \citep{levy2015live} and also record the off-by-one-accuracy (OBOA) over the entire dataset:
\begin{align}
  \text{MAE} &= \frac{1}{N} \sum_{i=1}^N \left|\widehat{c}_i - c_i\right| /c_i \\
  \text{OBOA} &= \frac{1}{N} \sum_{i=1}^N \big[ \left|\hat{c}_i - c_i \right| \leq 1 \big]
\end{align}
The mean-absolute error is preferred over the common mean-squared error as it is relative to the true count. To account for rounding errors and possible cycle cut-offs at both ends of the video, the off-by-one-accuracy is more suitable than the traditional accuracy.

%These evaluation metrics are not perfect as they highly penalize small count errors when the ground truth count is small. %\todo{Can we evaluate in sliding window fashion? We have the current cycle length over the video. We can compute the per-segment mean squared error. Then we overcome reducing the evaluation to a single number.}

%!TEX root = ms.tex

\subsection{Implementation Details}
\label{subsec:implementation-details}

\noindent \textbf{Optical Flow.} Our method takes two consecutive video frames as input and first estimates the motion using optical flow. As the quality of motion estimation may be important, we measure our method's sensitivity to three flow estimation methods. To evaluate a more traditional flow estimation method we choose TV-L$^1$ \citep{zach2007duality}. This variational based method is still competitive with more recent methods. Current state-of-the-art motion estimation methods all use convolutional neural networks for the purpose. We compare the deep learning based methods EpicFlow \citep{revaud2015epicflow} and FlowNet 2.0 \citep{ilg2017flownet}. Both deep networks are trained on large (synthetic) video datasets to estimate the motion in complex video. As default we use FlowNet 2.0. \\ % unless explicitly mentioned. \\

\begin{figure*}[t]
  \centering
  \includegraphics[trim={0.5cm 0.5cm 0 0 },clip,width=\textwidth]{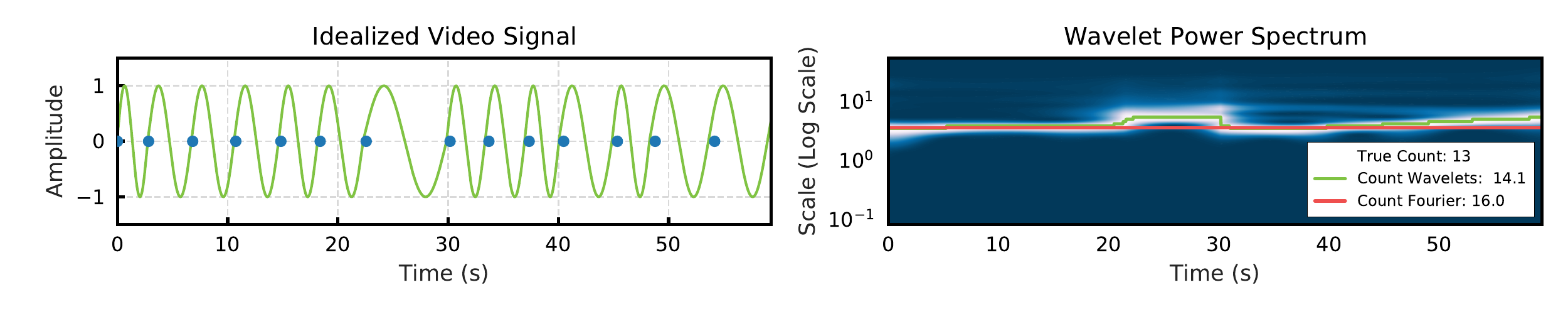}
  \caption{Idealized signal for a difficult non-stationary video displaying a violin player. The blue markers indicate the cycle bounds, manually annotated for each video in our \datasetname{} dataset. Note how the wavelet scalogram correctly exposes the rhythmic slowdown (around $20$ seconds). On the right, the green line corresponds to the local frequency predictions from the scalogram whereas the red (straight) line indicates the stationary Fourier-based frequency measurement. This demonstrates the effectiveness of wavelet analysis for optical non-stationary video signals.
  \label{fig:idealized-signals-with-scalogram}}
\end{figure*}

\noindent \textbf{Motion Segmentation.} Complex videos with background clutter or camera motion demand segmentation of the foreground motion prior to further analysis. Although our method directly performs localization from the densely computed wavelet power, we also evaluate with state-of-the-art motion segmentation methods. The fast video segmentation method of \cite{papazoglou2013fast} is chosen as classical approach and was also used in \cite{runia2018real}. This approach separates foreground objects from the background in a video by combining motion boundaries followed by segmentation refinement. We also evaluate the more recent deep learning based method of \cite{tokmakov2017}. The method trains a two-stream convolutional neural network with a long-short term memory (LSTM) module to capture the evolution over time. The network parameters are optimized using the large FlyingThings 3D dataset \citep{mayer2016large}. To refine the motion masks from the trained networks, a conditional random field is applied for refinement. For both methods we use the official implementations made available by the authors. While both methods generally attain excellent segmentations, we observed that segmentation fails completely for some more  difficult frames (either all or none pixels selected as foreground). To remedy incorrect segmentation masks we reuse the segmentation of the previous frame if the fraction of foreground pixels is less than $1\%$ of the entire frame. \\

\noindent \textbf{Differential Geometric Motion Maps.} To compute the motion maps we perform spatial filtering by first-order Gaussian kernels. The filtering is implemented in PyTorch and runs in large batches on the GPU to accelerate computation. Spatial convolution is performed with $\sigma = 4$ for all experiments. We also evaluated $\sigma = \{2,8,16\}$ but found only minor variation in performance. In practice, a combination of multiple spatial scales may produce best results. Once the spatial first-order derivatives $\nabla_x F_x, \nabla_y F_x, \nabla_x F_y$ and $\nabla_y F_y$ have been obtained through convolution, the differential motion maps are computed as specified in Section~\ref{subsec:methods-differential-motion-estimation}. \\

\noindent \textbf{Continuous Wavelet Transform.} We use the continuous wavelet filtering implementation as outlined in \cite{torrence1998practical}. In comparison to the previous version of our work, we now also perform temporal filtering on the GPU\footnote{\,\url{https://github.com/tomrunia/PyTorchWavelets}} resulting in a considerable speed-up. This enables us to apply the wavelet transform in large batches over all spatial locations in the video. As previously mentioned, we use a Morlet wavelet ($\omega_0 = 6$) with logarithmic scales ($\delta j = 0.125$, $s_0 = 2\delta t$). We limit the range of $J$ corresponding to a minimum of four repetitions by setting $s_{\min}$ and $s_{\max}$ accordingly in \eqref{eq:scale-distr1} and \eqref{eq:scale-distr2}. Depending on the video length, there are typically between $50$ and $60$ temporal scales levels. When compute budget is tight, computational efficiency can be improved by pruning the filter bank with scale selection, for example using the maximum response of a Laplacian filter \citep{lindeberg2017dense}. \\

\noindent \textbf{Repetition Counting.} The instantaneous frequency estimates are obtained from the dense wavelet power by pooling over the motion foreground mask. As detailed in Section~\ref{subsec:methods-repetition-counting}, the frequencies are integrated over time to arrive at a final repetition count. To remove frequency estimate outliers inconsistent with adjacent frames, we apply a median filter of $9$ timesteps (frames) to enforce local smoothness. This gives a slight improvement on both video datasets. The final Count predictions are not rounded, hence evaluation metrics may be slightly off due to incomplete cycles. \\

\noindent \textbf{Reimplementation of Baselines.} We compare our method against two existing works for repetition estimation. The method of \cite{pogalin2008visual} is chosen to represent the class of Fourier-based methods. Our reimplementation uses a more recent object tracker \citep{henriques2012tracker} but is identical otherwise. The tracker is initialized by manually drawing a box on the first frame. Converting the frequency to a count is trivial using the video length and frame rate. Additionally, we compare with the deep learning method of \cite{levy2015live} using their publicly available code and pretrained model without any modifications.

%!TEX root = ms.tex

%%%%%%%%%%%%%%%%%%%%%%%%%%%%%%%%%%%%%%%%%%%%%%%%%%%%%%%%%%%%%%%%%%%%%%%%%%%%%%%%
%%%%%%%%%%%%%%%%%%%%%%%%%%%%%%%%%%%%%%%%%%%%%%%%%%%%%%%%%%%%%%%%%%%%%%%%%%%%%%%%
%%%%%%%%%%%%%%%%%%%%%%%%%%%%%%%%%%%%%%%%%%%%%%%%%%%%%%%%%%%%%%%%%%%%%%%%%%%%%%%%
% EXPERIMENT: Wavelets versus Fourier

\subsection{Temporal Filtering: Fourier versus Wavelets}
\label{subsec:experiments-temporal-filtering}

\noindent \textbf{Setup.} The goal of our first experiment is to demonstrate the effectiveness of the continuous wavelet transform for counting repetitions in non-stationary signals. We compare the stationary Fourier-based periodogram with the time-scale representation given by the wavelet scalogram. To isolate the effect of frequency measurements, we generate idealized signals of the videos in our \datasetname{} dataset. Specifically, we fit sinusoidal signals through the individual cycle bounds for each video to obtain simple 1D waveforms representing the video. \autoref{fig:idealized-signals-with-scalogram} shows an idealized signal example and the corresponding wavelet spectrum with count predictions. To compare with the Fourier-based measurement, we compute the periodogram, detect the maximum frequency peak and convert the corresponding frequency to a count using the video's duration. This yields a repetition count prediction for both the stationary and non-stationary measurements that we evaluate over the entire dataset. \\

\begin{figure}[b!]
  \centering
  %\vspace{-1em}
  \includegraphics[trim={0.4cm 0 0.4cm 0},clip,width=\columnwidth]{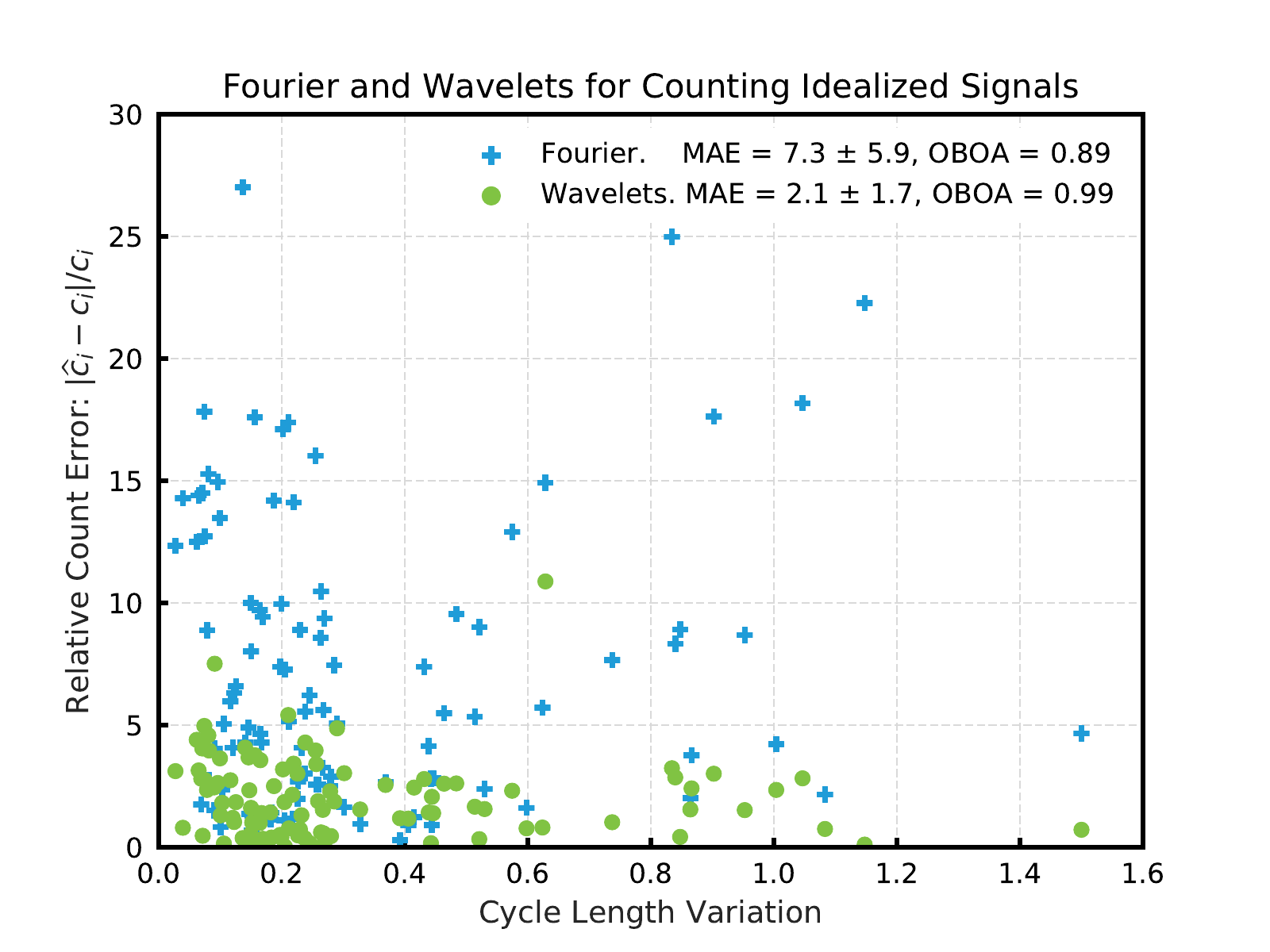}
  \caption{Fourier- versus wavelet-based repetition counting on idealized signals videos from the \datasetname{} dataset. Our wavelet-based method outperforms a Fourier-based baseline for $83$ out of $100$ videos. High cycle length variation results in notable error for Fourier measurements, whereas the time-localized wavelets are less sensitive to non-stationary repetition.
  \label{fig:scatter-fourier-wavelets-comparison}}
\end{figure}

\noindent \textbf{Results.} From the results in \autoref{fig:scatter-fourier-wavelets-comparison} it is clear that wavelet-based counting outperforms the periodogram on idealized signals. As expected, we observe that the Fourier-based measurements generally fail on videos with significant cycle length variation as they give a global frequency prediction. Wavelets naturally handle non-stationary repetition and are less sensitive to cycle length variability. We also tried adding a substantial amount of Gaussian noise ($\sigma = 0.5$) to the signals; this resulted in a minor negative effect on both methods (data not shown). This controlled experiment shows the effectiveness of wavelets for repetition estimation assuming a clear signal can be distilled from the videos.

%%%%%%%%%%%%%%%%%%%%%%%%%%%%%%%%%%%%%%%%%%%%%%%%%%%%%%%%%%%%%%%%%%%%%%%%%%%%%%%%
%%%%%%%%%%%%%%%%%%%%%%%%%%%%%%%%%%%%%%%%%%%%%%%%%%%%%%%%%%%%%%%%%%%%%%%%%%%%%%%%
%%%%%%%%%%%%%%%%%%%%%%%%%%%%%%%%%%%%%%%%%%%%%%%%%%%%%%%%%%%%%%%%%%%%%%%%%%%%%%%%
% EXPERIMENT: Viewpoint Relative to Motion

\subsection{Viewpoint Invariance}
\label{subsec:viewpoint-invariance}

\begin{figure*}

  \centering

  \begin{subfigure}{0.09\textwidth}
    \centering
    \includegraphics[width=\textwidth]{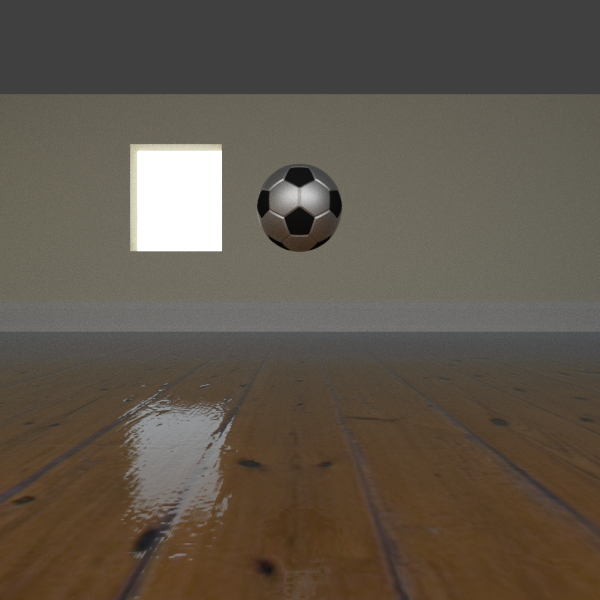}
  \end{subfigure}
  \hspace{0.1em}
  \begin{subfigure}{0.09\textwidth}
    \centering
    \includegraphics[width=\textwidth]{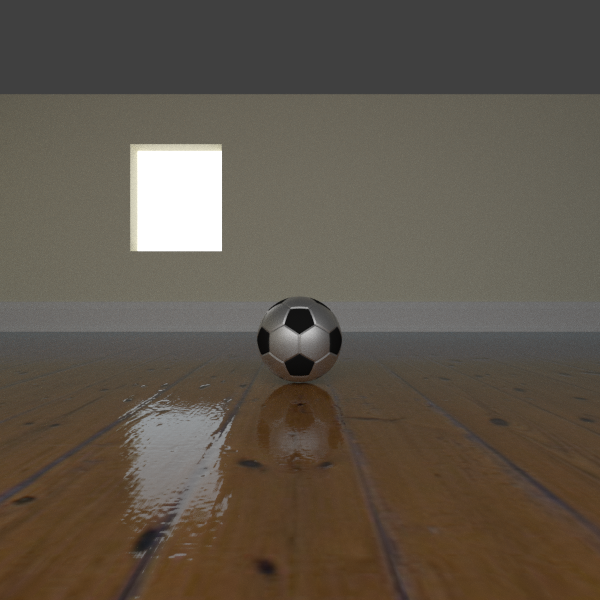}
  \end{subfigure}
  \hspace{0.1em}
  \begin{subfigure}{0.09\textwidth}
    \centering
    \includegraphics[width=\textwidth]{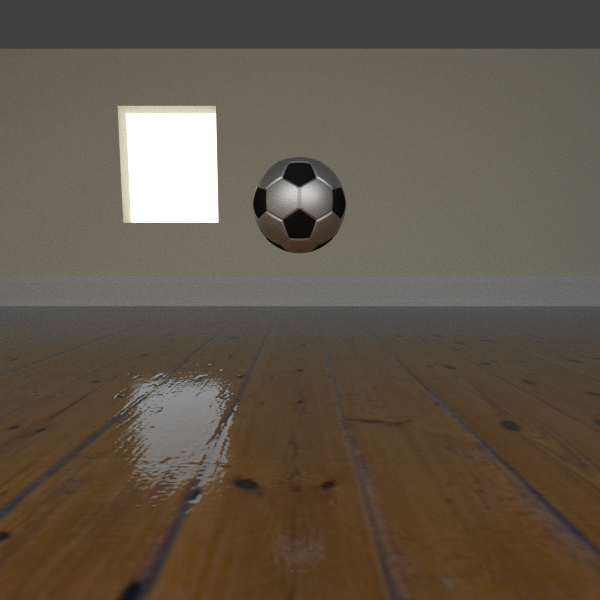}
  \end{subfigure}
  \hspace{0.1em}
  \begin{subfigure}{0.09\textwidth}
    \centering
    \includegraphics[width=\textwidth]{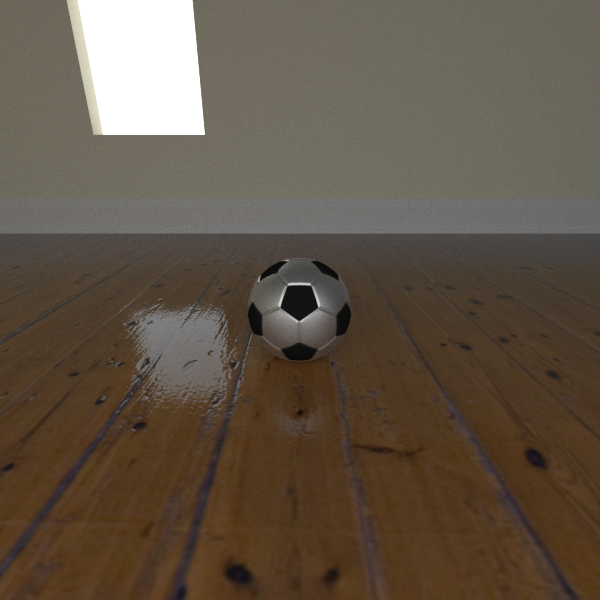}
  \end{subfigure}
  \hspace{0.1em}
  \begin{subfigure}{0.09\textwidth}
    \centering
    \includegraphics[width=\textwidth]{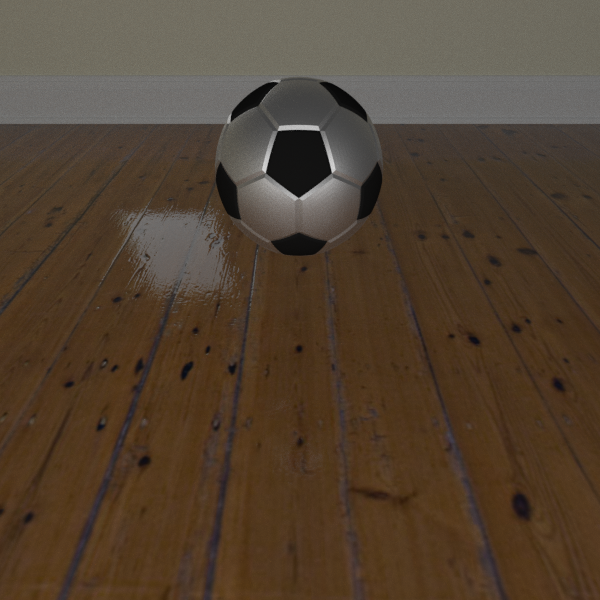}
  \end{subfigure}
  \hspace{0.1em}
  \begin{subfigure}{0.09\textwidth}
    \centering
    \includegraphics[width=\textwidth]{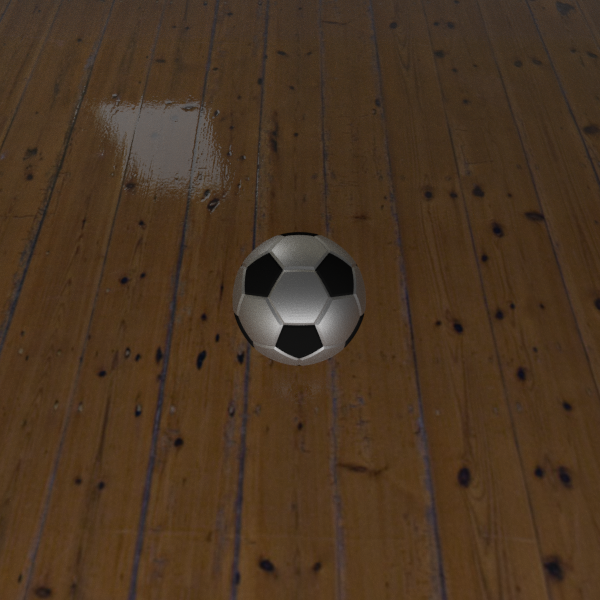}
  \end{subfigure}
  \hspace{0.1em}
  \begin{subfigure}{0.09\textwidth}
    \centering
    \includegraphics[width=\textwidth]{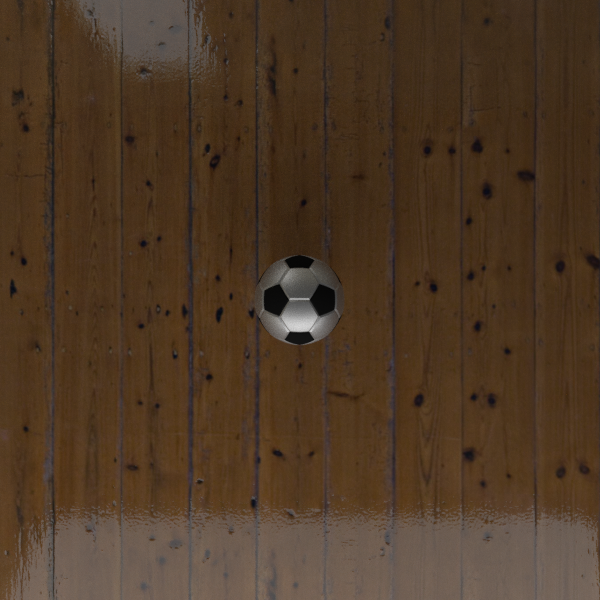}
  \end{subfigure}
  \hspace{0.1em}
  \begin{subfigure}{0.09\textwidth}
    \centering
    \includegraphics[width=\textwidth]{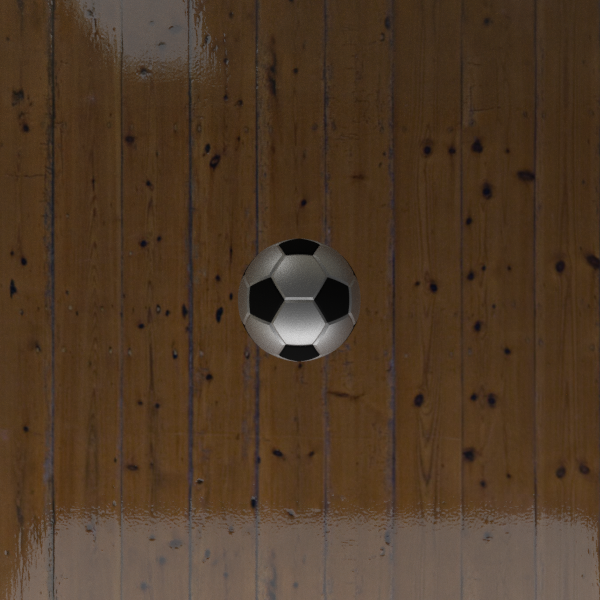}
  \end{subfigure}
  \hspace{0.1em}
  \begin{subfigure}{0.09\textwidth}
    \centering
    \includegraphics[width=\textwidth]{blender_viewpoint_experiment/0225}
  \end{subfigure}
  \hspace{0.1em}
  \begin{subfigure}{0.09\textwidth}
    \centering
    \includegraphics[width=\textwidth]{blender_viewpoint_experiment/0206}
  \end{subfigure}
  \\[1em]
  \begin{subfigure}{1.0\textwidth}
    \centering
    \includegraphics[trim={0.4cm 0.2cm 0.5cm 0},clip,width=\textwidth]{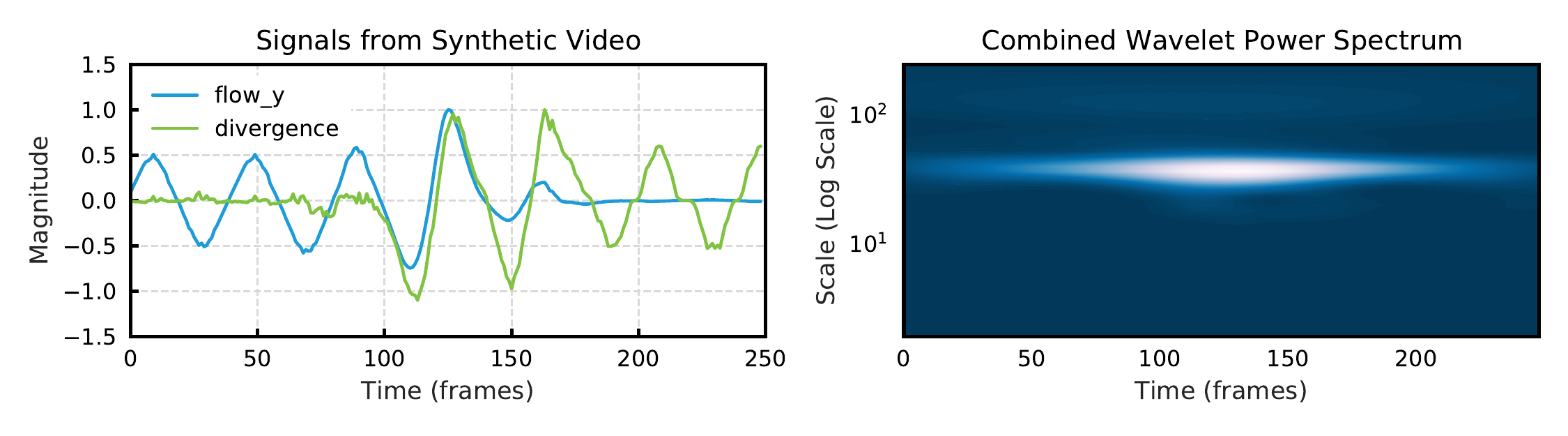}
  \end{subfigure}

  \caption{\emph{Top}: synthetized video sequence for a controlled experiment on the influence of viewpoint relative to the motion. This video clip shows a 3D modeled scene containing a bouncing ball. At the midpoint of the animation, the camera smoothly transitions from side view to frontal view. \emph{Bottom Left}: the time-varying magnitude of vertical flow and divergence measured over the foreground segmentation. Initially, the vertical flow is dominant and divergence is negligible. This reverses with the viewpoint transition. \emph{Bottom Right}: the combined wavelet spectrum of both signals. Notice the spectrum's invariance to viewpoint change as a result of wavelet power summation.}
  \label{fig:blender-viewpoint-experiment}

\end{figure*}

\noindent \textbf{Setup.} The theory of repetition considers two viewpoint extremes (\autoref{fig:classification-motion-types}). In this experiment we evaluate our method's ability to handle a continuous transition from one viewpoint extreme to the other. The designated mechanism for this is the use of multiple motion representations and the summation of their spectral power obtained from the continuous wavelet transform. To test this, we set-up a controlled experiment in which we synthesize a video clip from 3D modeled data in Blender. This enables full control over the object's motion and the viewpoint. Specifically, we choose to build a simple 3D scene containing a ball periodically bouncing on the floor as displayed in the top row of \autoref{fig:blender-viewpoint-experiment}. Initially, the camera captures the bouncing ball from the side view but after a number of full motion cycles, the camera smoothly transitions to frontal view (case $3$ to case $6$ in \autoref{fig:classification-motion-types}). We record the median-pooled vertical flow and divergence over the foreground region to obtain two time-varying signals. The spectral power for both signals is individually estimated using the continuous wavelet transform, after which we combine the power by summation. \\

\noindent \textbf{Results.} \autoref{fig:blender-viewpoint-experiment} plots the two median-pooled flow signals and their joint wavelet power obtained by summation. Initially, as the moving object is captured from the side view, vertical flow is best measurable. Upon the viewpoint transition, vertical flow vanishes while the divergent flow becomes dominant. As a result of the camera motion, the measurement of the spectral power for both individual signals will only give a strong response for either the first or second half of the video. However, the summation of the spectra gives a clear measurement over the complete video as is apparent from the combined wavelet power spectrum. This illustrates our method's ability to handle viewpoint changes by the combination of the wavelet power contained in multiple motion representations. By summation of the spectra, the best measurable motion representation will naturally give the largest contribution to the combined power. Therefore, this mechanism acts as a replacement of the global representation selection used in \citep{runia2018real} by dynamically leveraging information in all representations.

% Effectively, this mechanism dynamically selects the motion representation measurable best. This is different from our previous work \citep{runia2018real} where the single best representation was selected over the entire video, thus not suitable for dealing with significant viewpoint changes.

%%%%%%%%%%%%%%%%%%%%%%%%%%%%%%%%%%%%%%%%%%%%%%%%%%%%%%%%%%%%%%%%%%%%%%%%%%%%%%%%
%%%%%%%%%%%%%%%%%%%%%%%%%%%%%%%%%%%%%%%%%%%%%%%%%%%%%%%%%%%%%%%%%%%%%%%%%%%%%%%%
%%%%%%%%%%%%%%%%%%%%%%%%%%%%%%%%%%%%%%%%%%%%%%%%%%%%%%%%%%%%%%%%%%%%%%%%%%%%%%%%
% EXPERIMENT: Diversity in Motion Maps

\subsection{Diversity in Motion Maps}
\label{subsec:experiments-diversity-motion-differentials}

\noindent{\textbf{Setup.}} As wavelets prove to be effective for repetition estimation and multiple representations show value on a synthetic video, we now assess the value of a diversified video representation on real videos of our \datasetname{} dataset. We hypothesize that, due to the high variability in motion pattern and viewpoint, no single representation is powerful but their joint diversity is effective. To test this, we perform repetition counting over all individual motion maps listed in Eq.~\eqref{eq:six-motion-maps}. Instead of summing the wavelet power for all representations, we test the performance of the six motion representations individually. For each representation we densely compute te wavelet power and count the number of repetitions as outlined in the method's section. For a fair comparison, we exclude our motion segmentation mechanism based on wavelet power and instead use the motion segmentation proposed by \cite{papazoglou2013fast}. Again, we evaluate repetition counting on our \datasetname{} dataset. To obtain a lower-bound on the error, we also select the best representation per video in an oracle fashion. \\

\noindent{\textbf{Results.}} The results in \autoref{tab:quva-recurrence-individual-signals} reveal that for the wide variability of repetitive appearance there is no one size fits all solution. The individual motion maps are unable to handle the variety of repetitive motion appearances by themselves, resulting in poor count performance over the dataset. However, their joint diversity produces a good lower-bound by oracle selection of the most discriminative motion map. We notice the superiority of vertical flow $F_y$ as it performs best and is selected most often by the oracle. We explain this bias towards vertical flow by the observation that our dataset contains several sports videos in which the gravity is often used as opposing force.

\begin{table}
\centering
\caption{Value of diversity in six motion maps for videos from \datasetname{}. The last column denotes how often each signal is selected by the oracle. While the individual signals struggle to obtain good performance by themselves, exploiting their joint diversity is beneficial.}
\label{tab:quva-recurrence-individual-signals}
  \begin{tabular}{lrrcc}
  \toprule
   & MAE & OBOA & \# Selected\\
  \midrule
  % Using FastVideoSegment
  $\divergence$     & $77.8 \pm 90.8$ & $0.21$ & $10$ \\
  $\curl$           & $53.0 \pm 65.5$ & $0.32$ & $11$ \\
  $\nabla_x F_x$    & $58.1 \pm 63.5$ & $0.29$ & $15$ \\
  $\nabla_y F_y$    & $59.5 \pm 68.4$ & $0.31$ & $9$ \\
  $F_x$             & $49.6 \pm 48.0$ & $0.35$ & $25$ \\
  $F_y$             & $42.0 \pm 45.3$ & $0.43$ & $30$ \\
  \midrule
  Oracle Best       & $24.1 \pm 33.5$ & $0.63$ & $100$ \\
  % USING OUR MOTION SEGMENTATION:
  % \midrule
  % $\divergence$       & $81.3 \pm 97.2$ & $0.23$ & $6$ \\
  % $\curl$             & $44.1 \pm 49.3$ & $0.37$  & $13$ \\
  % $\nabla_x F_x$    & $62.4 \pm 66.7$ & $0.24$  & $9$ \\
  % $\nabla_y F_y$    & $58.8 \pm 70.3$ & $0.33$  & $9$ \\
  % $F_x$             & $43.9 \pm 47.0$ & $0.42$  & $26$ \\
  % $F_y$             & $26.7 \pm 38.5$ & $0.57$  & $37$ \\
  % \midrule
  % Oracle Best       & $14.4 \pm 28.3$ & $0.76$ & $100$ \\
  \bottomrule
  \end{tabular}
\end{table}

%%%%%%%%%%%%%%%%%%%%%%%%%%%%%%%%%%%%%%%%%%%%%%%%%%%%%%%%%%%%%%%%%%%%%%%%%%%%%%%%
%%%%%%%%%%%%%%%%%%%%%%%%%%%%%%%%%%%%%%%%%%%%%%%%%%%%%%%%%%%%%%%%%%%%%%%%%%%%%%%%
%%%%%%%%%%%%%%%%%%%%%%%%%%%%%%%%%%%%%%%%%%%%%%%%%%%%%%%%%%%%%%%%%%%%%%%%%%%%%%%%
% EXPERIMENT: Video Acceleration Sensitivity

\begin{figure*}
\centering
\begin{minipage}{.48\textwidth}
  \centering
  \includegraphics[width=\textwidth]{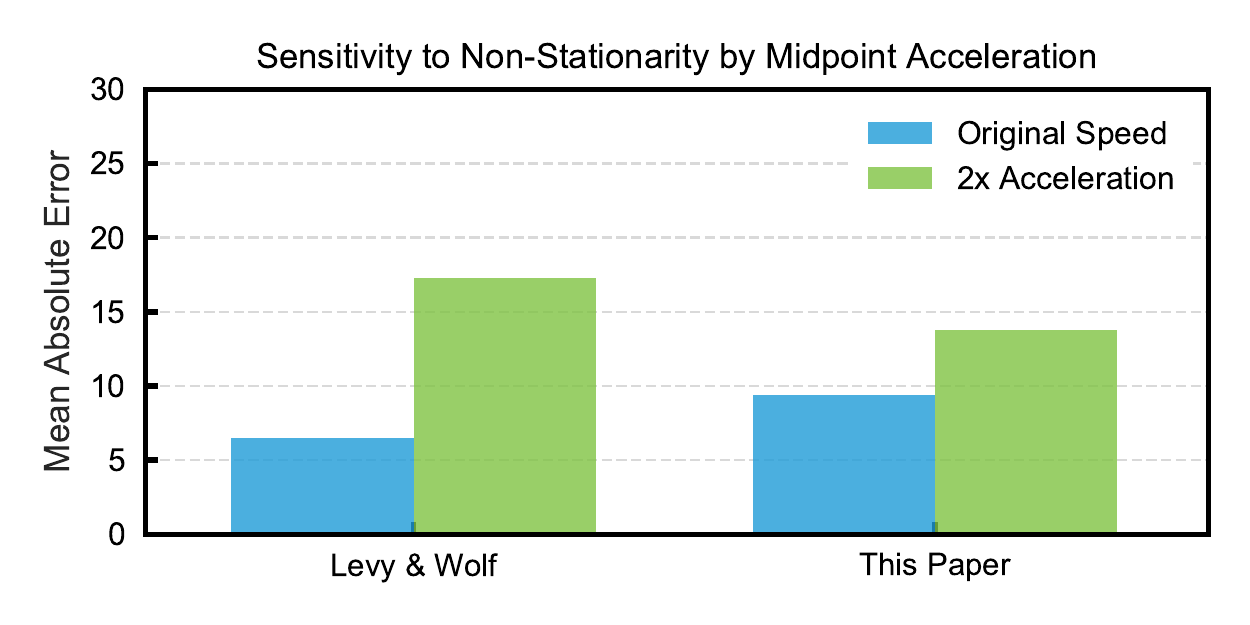}
  \caption{The effect of midpoint acceleration on the \ytsegments{} dataset. Our method increases $4.4$ in mean absolute error whereas the method of \cite{levy2015live} rises with $10.8$ points. The deep learning method has difficulty dealing with non-stationary acceleration, whereas our method is more robust due to the wavelet transform.}
  \label{fig:barchart-acceleration}
\end{minipage}
\quad\quad
\begin{minipage}{.48\textwidth}
  \centering
  \includegraphics[width=\textwidth]{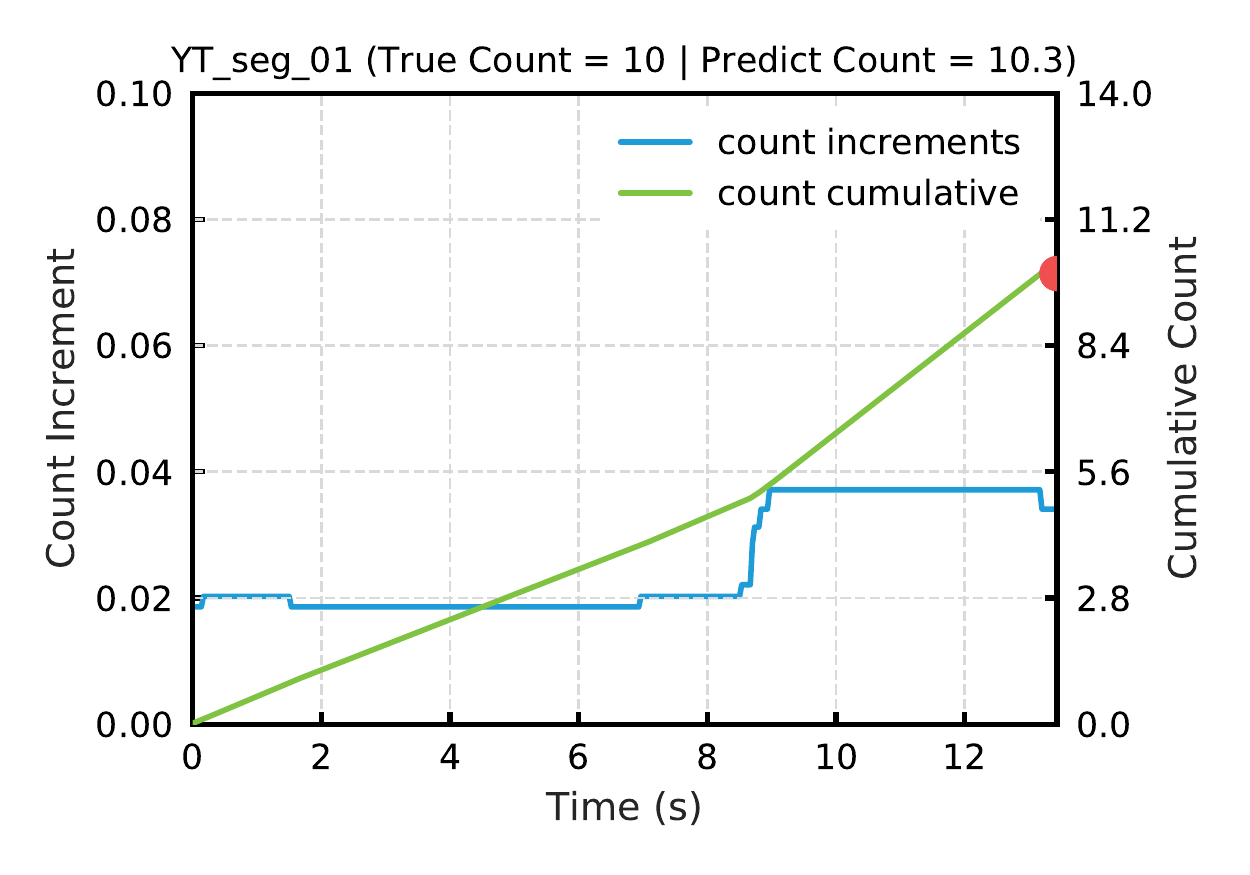}
  \caption{Count increments and cumulative count over time for the first video of \ytsegments{} with midpoint acceleration. The red marker on the right corresponds to the ground truth count. Note how the increase in speed around $9$~seconds is clearly reflected in the count increments.}
  \label{fig:count-increments-acceleration}
\end{minipage}
\vspace{-1em}
\end{figure*}

\subsection{Video Acceleration Sensitivity}
\label{subsec:experiments-halfway-acceleration}

\noindent \textbf{Setup.} In this experiment, we examine our method's sensitivity to acceleration by artificially speeding-up videos. Starting from the \ytsegments{} dataset, in which most videos exhibit strong periodic motion, we induce significant non-stationarity by artificially accelerating the videos halfway. More precisely, we modify the videos such that after the midpoint frame, the speed is increased by dropping every second frame. What follows are $100$ videos with a $2\times$ acceleration starting halfway. We compare against the deep learning method of \cite{levy2015live} which handles non-stationarity by running the period-predicting convolutional neural network in sliding-window fashion over the video. Fourier-based analysis was left out as it will inevitably fail on this task. \\

\noindent \textbf{Results.} The bar chart of \autoref{fig:barchart-acceleration} presents the mean absolute error in both original and accelerated setting. On their own dataset, the system of \cite{levy2015live} slightly outperforms our method. Acceleration reverses the results as our method suffers less and obtains a lower error on the accelerated videos. It reveals their sensitivity to acceleration, whereas our method deteriorates less. This shows the effectiveness of wavelets for dealing with non-stationarity in realistic videos. To illustrate how our method deals with midpoint acceleration, we also plot the count increments and cumulative counts throughout the video; see \autoref{fig:count-increments-acceleration}. As is evident from the plot, there is a distinct increase in count increments per timestep when upon enabling acceleration. This is observed for most videos in the dataset. This could be beneficial for detecting acceleration or temporal localization of transient phenomena in video.

%%%%%%%%%%%%%%%%%%%%%%%%%%%%%%%%%%%%%%%%%%%%%%%%%%%%%%%%%%%%
%%%%%%%%%%%%%%%%%%%%%%%%%%%%%%%%%%%%%%%%%%%%%%%%%%%%%%%%%%%%

\subsection{Motion Segmentation}
\label{subsec:experiments-repetitive-motion-segmentation}

\noindent \textbf{Setup.} In this experiment investigate the effectiveness of the motion segmentations obtained directly from the wavelet power for repetition estimation. We visually compare the motion segmentations and test whether replacing our localization mechanism with a state-of-the-art motion segmentation method improves repetition estimation performance. We keep the method identical except for the segmentation method to obtain a motion mask. In addition to our wavelet-based motion segmentation to obtain the discriminative motion mask we compare our method's performance without any localization (full-frame), the video segmentation method of \cite{papazoglou2013fast} and the deep learning approach of \cite{tokmakov2017}. \\

\begin{figure*}
  \centering

  \begin{subfigure}{0.32\textwidth}
    \centering
    \includegraphics[width=\textwidth]{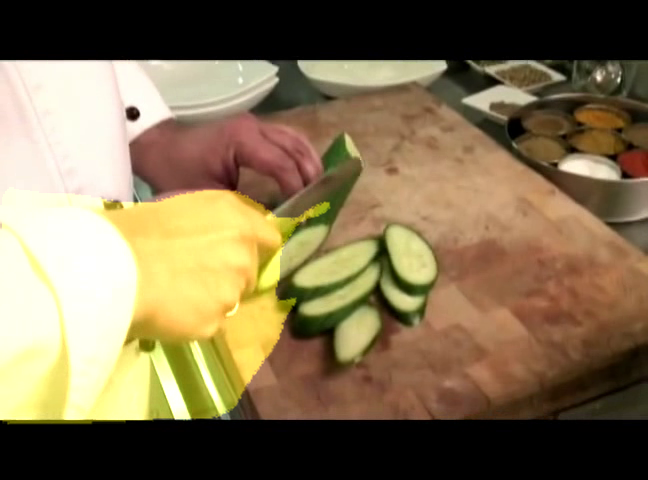}
  \end{subfigure}
  \,\,
  \begin{subfigure}{0.32\textwidth}
    \centering
    \includegraphics[width=\textwidth]{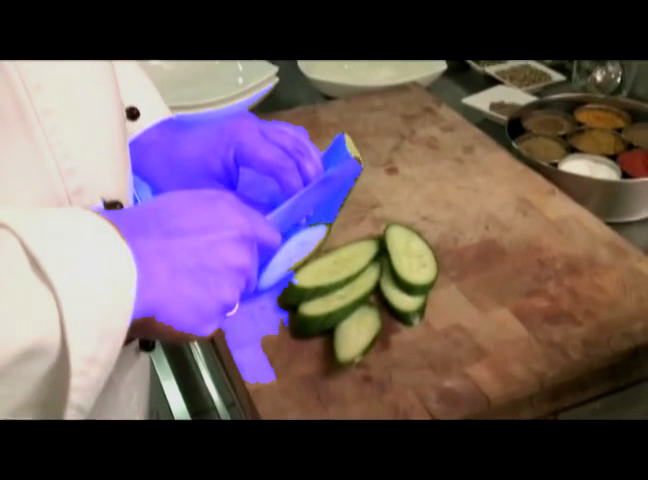}
  \end{subfigure}
  \,\,
  \begin{subfigure}{0.32\textwidth}
    \centering
    \includegraphics[width=\textwidth]{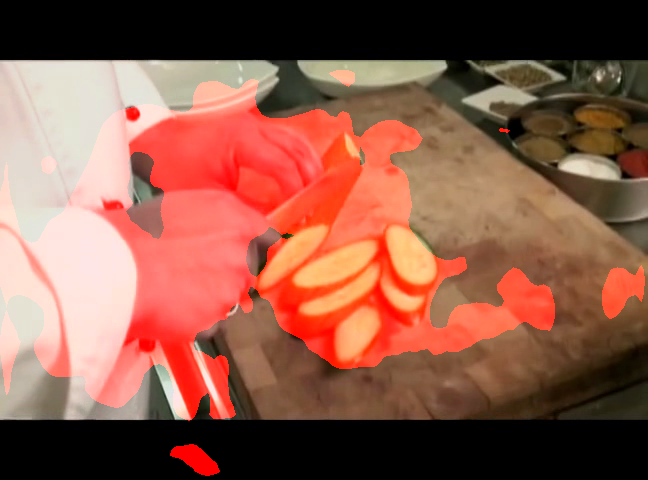}
  \end{subfigure}
  \\[0.5em]
  \begin{subfigure}{0.32\textwidth}
    \centering
    \includegraphics[width=\textwidth]{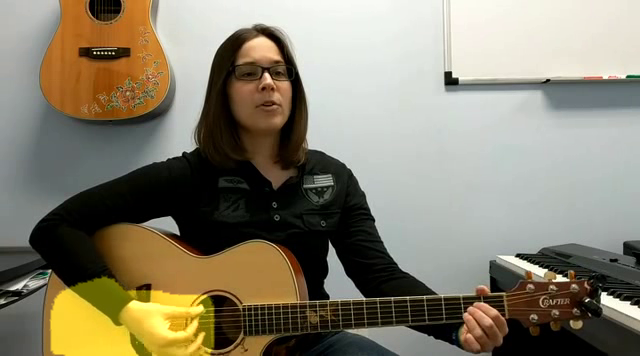}
  \end{subfigure}
  \,\,
  \begin{subfigure}{0.32\textwidth}
    \centering
    \includegraphics[width=\textwidth]{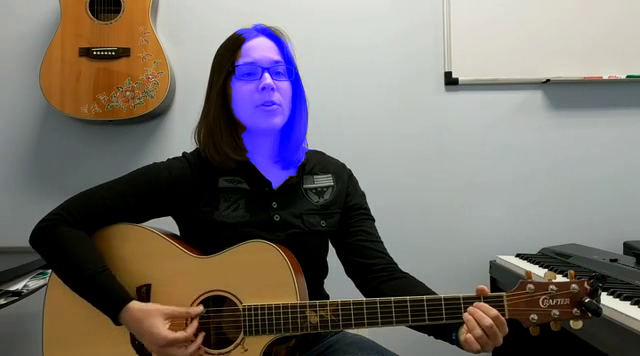}
  \end{subfigure}
  \,\,
  \begin{subfigure}{0.32\textwidth}
    \centering
    \includegraphics[width=\textwidth]{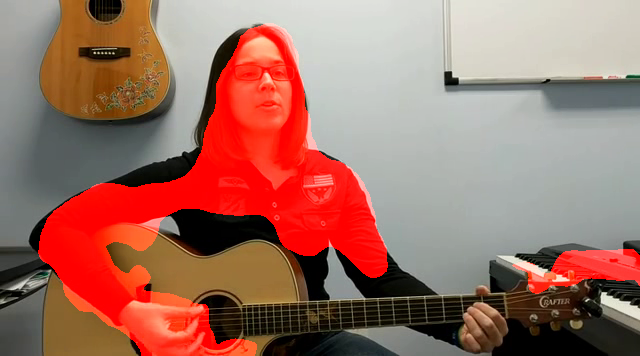}
  \end{subfigure}
  \\[0.5em]
  \begin{subfigure}{0.32\textwidth}
    \centering
    \includegraphics[width=\textwidth]{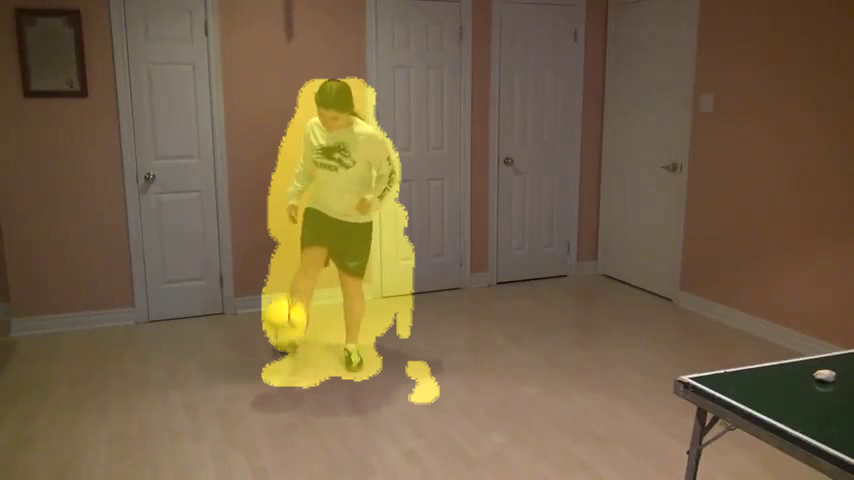}
  \end{subfigure}
  \,\,
  \begin{subfigure}{0.32\textwidth}
    \centering
    \includegraphics[width=\textwidth]{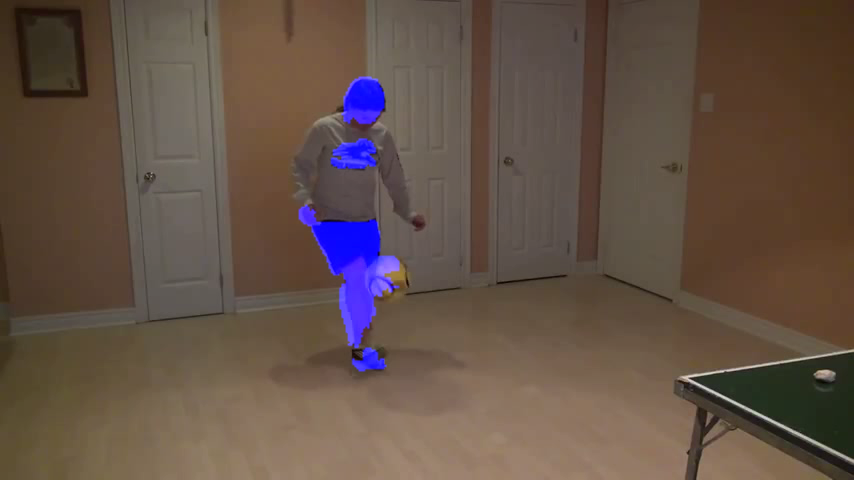}
  \end{subfigure}
  \,\,
  \begin{subfigure}{0.32\textwidth}
    \centering
    \includegraphics[width=\textwidth]{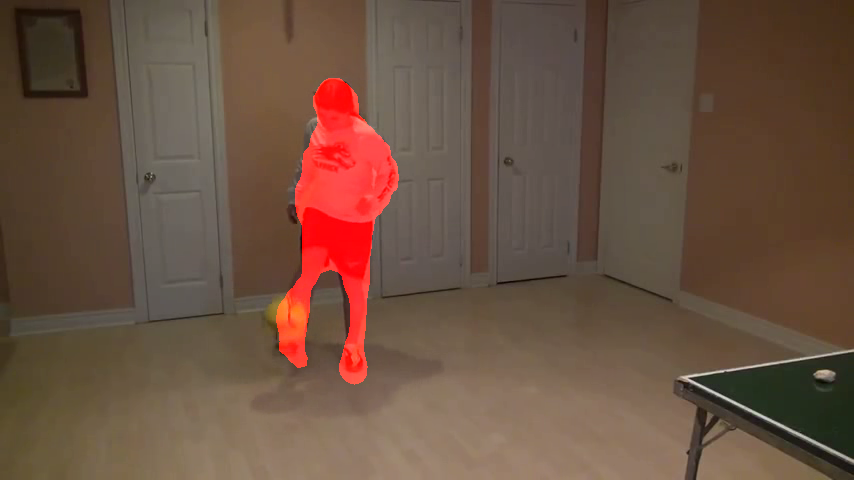}
  \end{subfigure}
  \\[0.5em]
  \begin{subfigure}{0.32\textwidth}
    \centering
    \includegraphics[width=\textwidth]{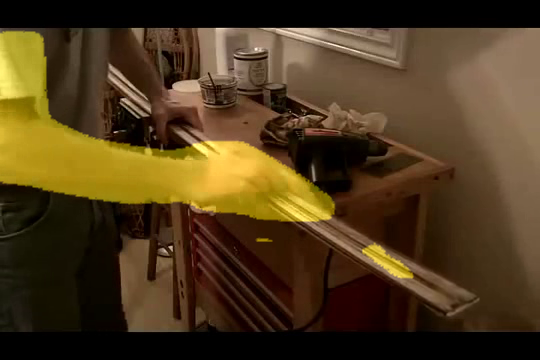}
  \end{subfigure}
  \,\,
  \begin{subfigure}{0.32\textwidth}
    \centering
    \includegraphics[width=\textwidth]{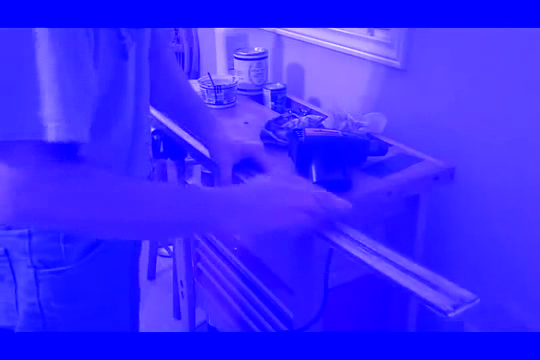}
  \end{subfigure}
  \,\,
  \begin{subfigure}{0.32\textwidth}
    \centering
    \includegraphics[width=\textwidth]{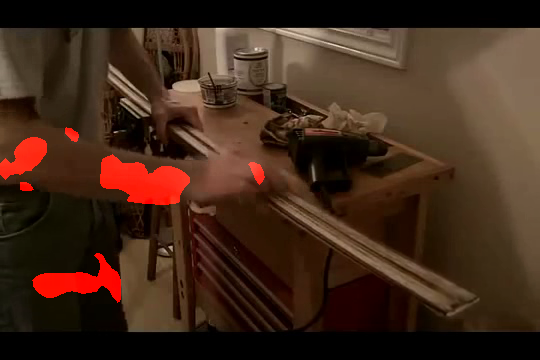}
  \end{subfigure}
  \\[0.5em]
  \begin{subfigure}{0.32\textwidth}
    \centering
    \includegraphics[width=\textwidth]{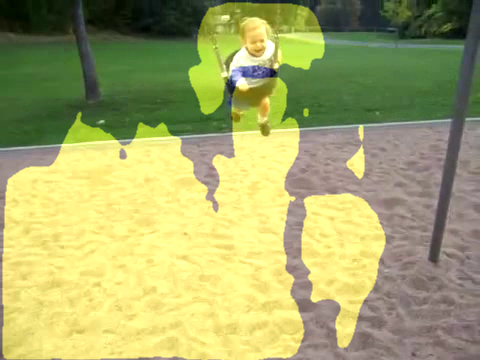}
  \end{subfigure}
  \,\,
  \begin{subfigure}{0.32\textwidth}
    \centering
    \includegraphics[width=\textwidth]{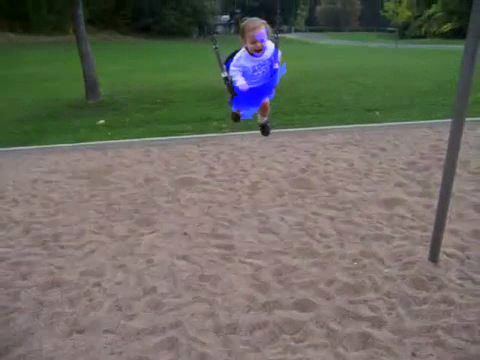}
  \end{subfigure}
  \,\,
  \begin{subfigure}{0.32\textwidth}
    \centering
    \includegraphics[width=\textwidth]{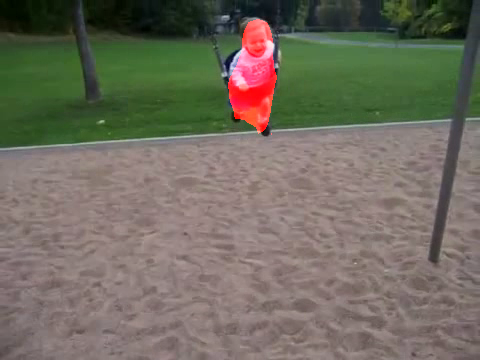}
  \end{subfigure}
  \\[0.5em]
  \begin{subfigure}{0.32\textwidth}
    \centering
    \includegraphics[width=\textwidth]{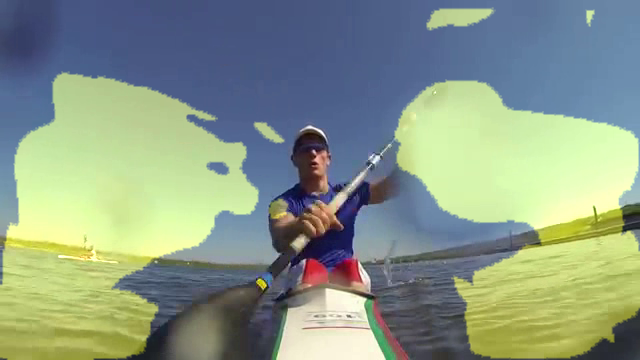}
    \caption{This paper}
  \end{subfigure}
  \,\,
  \begin{subfigure}{0.32\textwidth}
    \centering
    \includegraphics[width=\textwidth]{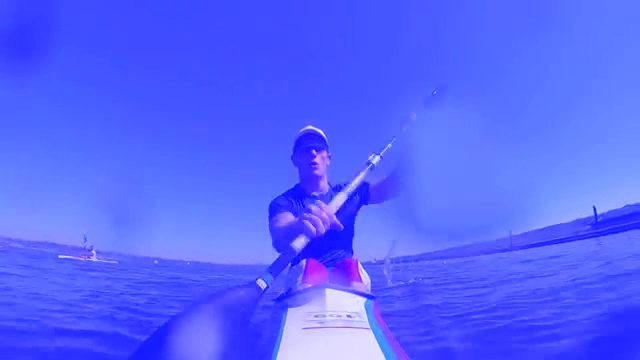}
    \caption{\cite{papazoglou2013fast}}
  \end{subfigure}
  \,\,
  \begin{subfigure}{0.32\textwidth}
    \centering
    \includegraphics[width=\textwidth]{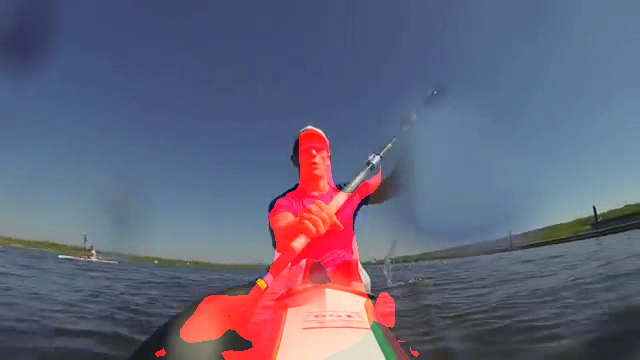}
    \caption{\cite{tokmakov2017}}
  \end{subfigure}

  \caption{Comparison of different motion segmentation masks. In most cases, our method succeeds to spatially segment the repetitive motion. In comparison to methods specifically devoted to the task of motion segmentation, our masks are less precise. However, as our numerical evaluation shows, our segmentation masks are more suitable for the task of repetition estimation. The most informative repetitive cues do not necessarily overlap with the foreground motion. In the last example, the regions through which the paddles moves produce the strongest repetitive response.} \label{fig:comparison-motion-segmentation-masks}
\end{figure*}

\noindent \textbf{Results.} We visually compare the three different motion segmentation methods in \autoref{fig:comparison-motion-segmentation-masks}. For most videos, our method is able to localize the repetitive motion. By all means, the state-of-the-art methods specifically devoted to foreground motion segmentation produce the visually best results. However, our intention is to obtain a motion mask best suitable for repetition estimation which not necessarily overlaps with the foreground motion. By thresholding the wavelet power maps, our method seems to emphasize on regions with most discriminative repetitive motion. This is best recognizable from the bottom two rows where the motion segmentation includes background regions that periodically change due to the motion. In \autoref{tab:motion-mask-comparison} we report quantitative results of our method with different motion segmentation methods. Our localization mechanism produces significantly better results than the existing motion segmentation methods. Partially, this might be explained by the temporal delay of the wavelet responses in comparison to the motion segmentation masks. For our method, this convincingly demonstrates that the segmentation directly obtained from the wavelet spectrum are more suitable than decoupled motion segmentation approaches.

% In comparison to the online method of \cite{levy2015live}, our method is unable not fully live as the wavelets are non-causal filters. In practice, this means that the filter responses are modestly delayed. For future work, we plan to perform time-causal filtering as proposed by \cite{lindeberg2016time} to provide a true online approach for live repetition estimation.

%%%%%%%%%%%%%%%%%%%%%%%%%%%%%%%%%%%%%%%%%%%%%%%%%%%%%%%%%%%%%%%%%%%%%%%%%%%%%%%%
%%%%%%%%%%%%%%%%%%%%%%%%%%%%%%%%%%%%%%%%%%%%%%%%%%%%%%%%%%%%%%%%%%%%%%%%%%%%%%%%

\begin{table*}
    \centering
    \caption{Repetition counting results of our method with different motion segmentation mechanism. While the state-of-the-art motion segmentation methods produce visually excellent results, their segmentations are suboptimal for the task of repetition estimation. This is expected as the most discriminative repetitive cues are not always contained in the foreground motion. See \autoref{fig:comparison-motion-segmentation-masks} for a visual comparison of segmentation masks.}
    \label{tab:motion-mask-comparison}
    \begin{tabular}{lrrrrrr}
        \toprule
        & \multicolumn{2}{c}{\ytsegmentsbold{}} & \multicolumn{2}{c}{\datasetnamebold{}} \\
        \cmidrule{2-3} \cmidrule{4-5}
        Motion segmentation method & MAE $\downarrow$ & OBOA $\uparrow$ & MAE $\downarrow$ & OBOA $\uparrow$ \\
        \midrule
        Full-frame    & $46.0 \pm 67.2$ & $0.28$ & $60.8 \pm 49.4$ & $0.22$ \\
        \cite{papazoglou2013fast} & $13.1 \pm 20.3$ & $0.78$ & $42.6 \pm 49.2$ & $0.44$ \\
        \cite{tokmakov2017} & $21.6 \pm 57.2$ & $0.76$ & $38.9 \pm 39.2$ & $0.42$ \\
        Differential geometry (this paper) & $\mathbf{9.4 \pm 17.4}$ & $\mathbf{0.89}$ & $\mathbf{26.1 \pm 39.6}$ & $\mathbf{0.62}$ \\
        \bottomrule
    \end{tabular}
    \vspace{0.3cm}
\end{table*}

%%%%%%%%%%%%%%%%%%%%%%%%%%%%%%%%%%%%%%%%%%%%%%%%%%%%%%%%%%%%%%%%%%%%%%%%%%%%%%%%
%%%%%%%%%%%%%%%%%%%%%%%%%%%%%%%%%%%%%%%%%%%%%%%%%%%%%%%%%%%%%%%%%%%%%%%%%%%%%%%%
%%%%%%%%%%%%%%%%%%%%%%%%%%%%%%%%%%%%%%%%%%%%%%%%%%%%%%%%%%%%%%%%%%%%%%%%%%%%%%%%
% EXPERIMENT: Comparison to State-of-the-Art

\subsection{Comparison to the State-of-the-Art}
\label{subsec:experiments-repetition-counting}

\noindent \textbf{Setup.} In this experiment, we perform a full comparison on the task of repetition counting for both video datasets. We compare against the Fourier-based method of \cite{pogalin2008visual} and the deep learning approach of \mbox{\cite{levy2015live}}. \\ %Our method essentially requires only one hyper-parameter: the spatial scale for Gaussian filtering, which was fixed to $\sigma = 4$ throughout the paper. \\ %Spatial multiscale filtering could improve results, but this was not tested. The distribution of temporal scales is automatically initialized by equations \eqref{eq:scale-distr1} and \eqref{eq:scale-distr2}. \\

\begin{table*}
    \centering
    \caption{Comparison with the state-of-the-art on repetition counting for the \ytsegments{} and our \datasetname{} dataset. The deep learning-based method of \cite{levy2015live} achieves good results on their own dataset of relatively clean videos. On our more realistic and challenging dataset, the current method improves considerably over the existing approaches. In comparison to our previous work, our method segments the repetitive motion directly rather than relying on decoupled motion segmentation.} %\emph{EF} corresponds to EpicFlow and \emph{FN} corresponds to FlowNet2. \todo{mention that we repeat our CVPR experiments with FlowNet2 (slight higher numbers)}}
    \label{tab:count-evaluation-results}
    \small
    \begin{tabular}{lrrrrrr}
        \toprule
        & \multicolumn{2}{c}{\ytsegmentsbold{}} & \multicolumn{2}{c}{\datasetnamebold{}} \\
        \cmidrule{2-3} \cmidrule{4-5}
        & MAE $\downarrow$ & OBOA $\uparrow$ & MAE $\downarrow$ & OBOA $\uparrow$ \\
        \midrule
        \cite{pogalin2008visual} & $21.9 \pm 30.1$ & $0.68$ & $38.5 \pm 37.6$ & $0.49$ \\
        \cite{levy2015live}       & $\mathbf{6.5 \pm \phantom{0}9.2}$ & $\mathbf{0.90}$ & $48.2 \pm 61.5$ & $0.45$ \\
        %\cite{runia2018real} & $10.3 \pm 19.8$ & $0.89$ & $\mathbf{23.2 \pm 34.4}$ & $0.62$ \\
        %\cite{runia2018real} (FN) & $11.7 \pm 22.6$ & $0.76$ & $\mathbf{19.5 \pm 23.7}$ & $0.59$ \\ % <== Results using FlowNet2 / mixed-pooling
        This paper & $9.4 \pm 17.4$ & $0.89$ & $\mathbf{26.1 \pm 39.6}$ & $\mathbf{0.62}$ \\
        \bottomrule
    \end{tabular}
\end{table*}

\noindent \textbf{Results.} The full count evaluation is presented in \autoref{tab:count-evaluation-results}. On their own \ytsegments{} dataset, the method of \cite{levy2015live} performs best with an MAE of $6.5$, where our method achieves a comparable error of $9.4$ and near-identical off-by-one accuracy. Despite the stationary nature of most videos in this dataset, the Fourier-based approach of \cite{pogalin2008visual} performs unfavorably compared to all other methods. 

The results change dramatically when considering our challenging \datasetname{} dataset; notably the deep learning approach of \cite{levy2015live} now performs the worst, with an MAE of $48.2$. This could possibly be explained by the fact that their network only considers four motion types during training or the convolutional network's fixed temporal input dimension posing a constraint on the effective motion periods (ranging from $0.2$ to $2.33$ seconds). Dealing with motion periods outside of this range most likely requires retraining the network. The Fourier-based method of \cite{pogalin2008visual} scores an MAE of $38.5$, whereas we obtain an average error of $26.1$. On the \ytsegments{} dataset our simplified method slightly improves over the MAE of $10.3 \pm 19.8$ reported in \citep{runia2018real}, while giving comparable results to previously reported MAE of $23.2 \pm 34.4$ on the \datasetname{} dataset. The Fourier-based and deep learning-based approaches are unable to effectively handle the increased non-stationarity and motion complexity found in our challenging video dataset. The method proposed here improves the ability to handle such difficult videos without relying on explicit motion segmentation methods.

We also report the repetition count results using \mbox{TV-L$^1$} \citep{zach2007duality} and EpicFlow \citep{revaud2015epicflow} to investigate our method's sensitivity to optical flow quality. The results in \autoref{tab:flow-comparison} show the robustness to different flow methods as the algorithm of choice has limited effect on the count performance for both datasets. 

To gain a better understanding of our method's characteristics we study success and failure cases. We observe that our wavelet-based motion segmentation struggles with scenes containing water (\eg \autoref{fig:dataset-examples}, bottom row) or other dynamic texture. The rippling water produces visual repetitive dynamics resulting in a strong wavelet response over its entire surface. Consequently, motion segmentation by mean-thresholding of the spectral power will fail inevitably; and subsequent measurements over the foreground motion mask will be incorrect as well. For such videos, we observe an enormous over-count as the frequency estimates correspond to the high-frequent rippling water. The error associated with these videos explains the limited improvement over our previous method \citep{runia2018real} which relied on \cite{papazoglou2013fast} for motion segmentation, being less prone to such segmentation failures. 

We also observe that all methods make a common mistake: over-counting videos with a factor of two. The similarity in these videos is that one full cycle contains the exact same motion first with one arm (or leg) followed by the other (\eg walking lunges or swimming front-crawl). As the perceived motion is almost identical for both limbs, the estimated temporal dynamics are twice as fast. Again, the significant over-estimate of the motion frequency produces a large count error for all methods. Solving this problem is not easy, as current repetition estimates in those cases are essentially also a correct prediction; however, the human annotators define salient motion as a full cycle with both limbs. 

% We observe our method to perform well over the entire range of temporal periods contained on our dataset. This is explained by our dense temporal filter bank with logarithmic distribution of scales. The system of \cite{levy2015live} is trained to handle motion periods ranging from $0.2$ to $2.33$ seconds. Dealing with motion periods outside of this range requires most likely requires retraining of the network. As our method requires no learning, there is no limit on the cycle length range. We praise \cite{levy2015live} for the capability of true live counting; the repetitive count of our method is slightly delayed due to the non time-causal nature of wavelets. 

\begin{table}[b!]
    \centering
    \caption{Sensitivity of our method with respect to different optical flow methods. We report repetition counting results over both datasets. Only slight variation in the performance is observed, demonstrating our method's robustness to optical flow quality.}
    \label{tab:flow-comparison}
    \begin{tabular}{lrrrrrr}
        \toprule
        & \multicolumn{2}{c}{\ytsegmentsbold{}} & \multicolumn{2}{c}{\datasetnamebold{}} \\
        \cmidrule{2-3} \cmidrule{4-5}
        & MAE $\downarrow$ & OBOA $\uparrow$ & MAE $\downarrow$ & OBOA $\uparrow$ \\
        \midrule
        TV-L$^1$ & $9.8 \pm 17.9$ & $0.89$ & $26.5 \pm 67.5$ & $\mathbf{0.67}$ \\
        EpicFlow & $9.7 \pm 17.9$ & $0.88$ & $30.8 \pm 38.2$ & $0.55$ \\
        FlowNet 2.0 & $\mathbf{9.4 \pm 17.4}$ & $\mathbf{0.89}$ & $\mathbf{26.1 \pm 39.6}$ & $0.62$ \\
        \bottomrule
    \end{tabular}
\end{table}

%!TEX root = ms.tex

\section{Conclusion}
\label{sec:conclusion}

% Arnold: Conclusion:
% + Beperkt tot 1 paragraaf met 1 conclusie per novelty.
% + De laatste alinea kan eruit. Als het erin moet dan 1 zin in related work er ergens bij. Niet defensief eindigen.

We have categorized 3D intrinsic periodic motion as translation, rotation or expansion depending on the first-order differential decomposition of the motion field. Additionally, we distinguish three periodic motion continuities: constant, intermittent and oscillatory motion. For the 2D perception of 3D periodicity, the camera will be somewhere in the continuous range between two viewpoint extremes. What follows are $18$ fundamentally different cases of repetitive motion appearance in 2D. The practical challenges associated with repetition estimation are the wide variety in motion appearance, non-stationary temporal dynamics and camera motion. Our method addresses all these challenges by computing a diversified motion representation, employing the continuous wavelet transform and combining the power spectra of all representations to support viewpoint invariance. Whereas related work explicitly localizes the foreground motion, our method performs repetitive motion segmentation directly from the wavelet power maps resulting in a simplified approach. We verify our claims by improving the state-of-the-art on the task of repetition counting on our challenging new video dataset. The method requires no training and requires only a minimum number of hyper-parameters which are fixed throughout the paper. We envision applications beyond repetition estimation as the wavelet power and scale maps can support localization of low- and high-frequency regions suitable for region pruning or action classification.

%\input{07_appendices}  % before bib @ IJCV

%%%%%%%%%%%%%%%%%%%%%%%%%%%%%%%%%%%%%%%%%%%%%%%%%%%%%%%%%%%%%%%%%%%%%%%%%%%%%
%%%%%%%%%%%%%%%%%%%%%%%%%%%%%%%%%%%%%%%%%%%%%%%%%%%%%%%%%%%%%%%%%%%%%%%%%%%%%
%%%%%%%%%%%%%%%%%%%%%%%%%%%%%%%%%%%%%%%%%%%%%%%%%%%%%%%%%%%%%%%%%%%%%%%%%%%%%

\bibliographystyle{spbasic}   % basic style, author-year citations
\bibliography{bibliography}   % name your BibTeX data base}

\end{document}